\documentclass[times,twocolumn,final,authoryear]{elsarticle}

\usepackage{ycviu}
\usepackage{framed,multirow}

\usepackage{amssymb}
\usepackage{amsmath}
\usepackage{latexsym}

\usepackage{url}
\usepackage[table]{xcolor}
\definecolor{newcolor}{rgb}{.8,.349,.1}

\usepackage{subfig}
\usepackage{siunitx}
\usepackage{tikz}
\usepackage{booktabs}
\usepackage{wrapfig}

\newcommand{\figref}[1]{Fig.~\ref{#1}}

\newcommand{\secref}[1]{Section~\ref{#1}}
\renewcommand{\eqref}[1]{Equation~(\ref{#1})}

\definecolor{mred}{HTML}{A00000}
\definecolor{mgreen}{HTML}{00A000}
\definecolor{mblue}{HTML}{5060d0}

\def\pure{\emph{PuRe}}
\def\eyerectoo{\emph{EyeRecToo}}

\def\els{\emph{ElSe}}
\def\exc{\emph{ExCuSe}}
\def\swi{\emph{\'{S}wirski}}
\def\pup{\emph{PupilNet}}
\def\lpw{\emph{LPW}}
\def\clo{\emph{Closed-Eyes}}
\def\vo{\emph{Vera-Olmos}}
\def\an{$A_1$}
\def\ap{$A_2$}
\def\mtbf{\mathit{MTBF}}
\def\mttr{\mathit{MTTR}}

\def\tp{$\mathit{TP}$}
\def\fp{$\mathit{FP}$}
\def\tn{$\mathit{TN}$}
\def\fn{$\mathit{FN}$}

\def\ctp{$\mathit{CTP}$}
\def\itp{$\mathit{ITP}$}

\DeclareSIUnit\px{px}

\def\cpp{{C\nolinebreak[4]\hspace{-.05em}\raisebox{.4ex}{\scriptsize\bf ++}}}

\newcommand{\ds}[1]{\texttt{#1}}

\journal{Computer Vision and Image Understanding}

\begin{document}

\begin{frontmatter}

\title{PuRe: Robust pupil detection for real-time pervasive eye tracking }

\author[1]{Thiago \snm{Santini}\corref{cor1}\fnref{fn1}}
\cortext[cor1]{Corresponding author:}
\author[1]{Wolfgang \snm{Fuhl}\fnref{fn1}}
\fntext[fn1]{Authors contributed equally and should be considered co-first authors.}
\ead{thiago.santini@uni-tuebingen.de}
\author[1]{Enkelejda \snm{Kasneci}}

\address[1]{University of T\"ubingen, Sand 14, T\"ubingen -- 72076, Germany}

\received{1 May 2013}
\finalform{10 May 2013}
\accepted{13 May 2013}
\availableonline{15 May 2013}
\communicated{S. Sarkar}

\begin{abstract}

Real-time, accurate, and robust pupil detection is an essential prerequisite to
enable pervasive eye-tracking and its applications -- e.g., gaze-based human
computer interaction, health monitoring, foveated rendering, and advanced
driver assistance.
However, automated pupil detection has proved to be an intricate task in
real-world scenarios due to a large mixture of challenges such as quickly
changing illumination and occlusions.
In this paper, we introduce the \emph{\underline{Pu}pil
\underline{Re}constructor} (\pure), a method for pupil detection in pervasive
scenarios based on a novel edge segment selection and conditional segment
combination schemes; the method  also includes a confidence measure for the
detected pupil.
The proposed method was evaluated on over 316,000 images acquired with four
distinct head-mounted eye tracking devices.
Results show a pupil detection rate improvement of over 10 percentage points
w.r.t. state-of-the-art algorithms in the two most challenging data
sets (6.46 for all data sets), further pushing the envelope for pupil detection.
Moreover, we advance the evaluation protocol of pupil detection algorithms by
also considering eye images in which pupils are not present.
In this aspect, \pure{} improved \emph{precision} and \emph{specificity} w.r.t.
state-of-the-art algorithms by 25.05 and 10.94 percentage points, respectively,
demonstrating the meaningfulness of \pure's confidence measure.
\pure{} operates in real-time for modern eye trackers (at 120 \emph{fps}).

\end{abstract}


\end{frontmatter}

\section{Introduction}
\label{sec:introduction}

Head-mounted video-based eye trackers are becoming increasingly more accessible
and prevalent.
For instance, such eye trackers are now available as low-cost devices (e.g.,
~\cite{pupillabs2017}) or integrated into wearables such as Google Glasses,
Microsoft Hololens, and the Oculus
Rift~\citep{raffle2015heads,microsoft2017,oculus2017}.
As a consequence, eye trackers are no longer constrained to their origins as
research instruments but are developing into fully fledged pervasive devices.
Therefore, guaranteeing that these devices are able to seamlessly operate in
\emph{out-of-the-lab} scenarios is not only pertinent to the research of human
perception, but also to enable further applications such as pervasive gaze-based
human-computer interaction~\citep{bulling2010toward}, health
monitoring~\citep{vidal2012wearable}, foveated
rendering~\citep{guenter2012foveated}, and conditionally automated
driving~\citep{braunagel2017ready}.

\begin{figure}[h]
	\begin{center}
		\subfloat[][\label{a}]{
			\includegraphics[width=.23\columnwidth]{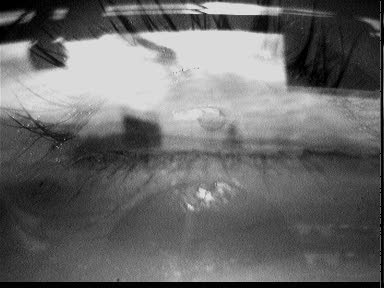}
			\label{fig:reflection}
		}
		\subfloat[][\label{b}]{
			\includegraphics[width=.23\columnwidth]{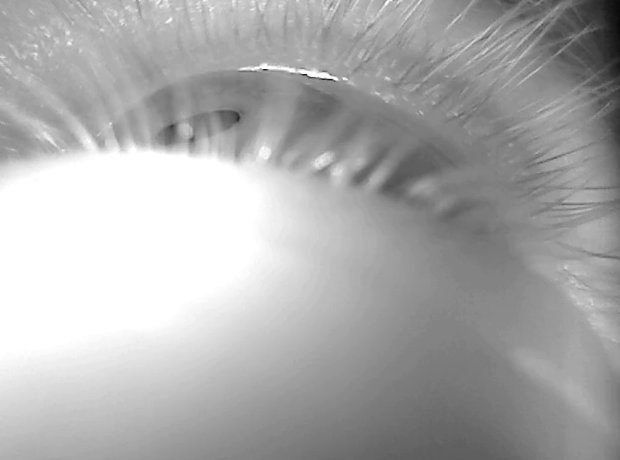}
			\label{fig:occlusion}
		}
		\subfloat[][\label{c}]{
			\includegraphics[width=.23\columnwidth]{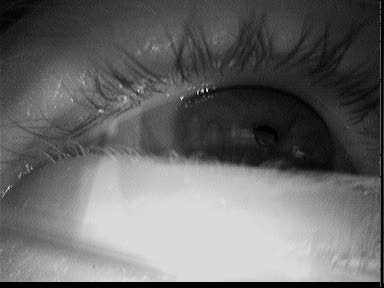}
			\label{fig:dark}
		}
		\subfloat[][\label{d}]{
			\includegraphics[width=.23\columnwidth]{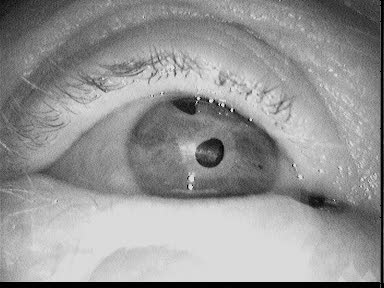}
			\label{fig:physiological}
		}
		\caption{
			Representative images of pupil detection challenges in real-world
			scenarios: (a) reflections, (b) occlusions, (c) complex
			illuminations, and (d) physiological irregularities.
		}
	\end{center}
\end{figure}

Pupil detection is the fundamental layer in the eye-tracking stack since most
other layers rely on the signal generated by this layer -- e.g., for gaze
estimation~\citep{morimoto2005eye}, and automatic identification of eye
movements~\citep{santini2016bayesian}.
Thus, errors in the pupil detection layer propagate to other layers,
systematically degrading eye-tracking performance.
Unfortunately, robust real-time pupil detection in natural environments has
remained an elusive challenge.
This elusiveness is evidenced by several reports of difficulties and low pupil
detection rates in natural environments such as
driving~\citep{schmidt2017eye,wood2017night,kubler2015driving,trosterer2014eye,kasneci2013towards,chu2010effect,liu2002real},
shopping~\citep{kasneci2014homonymous},
walking~\citep{sugano2015self,foulsham2011and}, and in an operating
room~\citep{tien2015differences}.
These difficulties in pupil detection stems from multiple factors; for instance,
reflections~(\figref{fig:reflection}), occlusions~(\figref{fig:occlusion}),
complex illuminations~(\figref{fig:dark}), and physiological
irregularities~(\figref{fig:physiological})~\citep{fuhl2016pupil,hansen2007improved,hansen2005eye,zhu2005robust}.

In this paper, we introduce the \emph{\underline{Pu}pil
\underline{Re}constructor} (\pure), a method for pupil detection in pervasive
scenarios based on a novel edge segment selection and conditional segment
combination schemes; the proposed method also includes a meaningful confidence
measure for the detected pupil.
Previous work in the field of pupil detection is presented in
\secref{sec:related}, and the proposed method is described in
\secref{sec:method}.
Previous work usually focuses on evaluating pupil detection algorithms based
solely on the detection rate. Similarly, we contrast the proposed method to
previous work in \secref{sec:detectionrate}.
Moreover, we go one step further and introduce novel metrics to evaluate
these algorithms in terms of incorrect pupil detection rates
(\secref{sec:eval-conf}) as well as dynamic signal properties
(\secref{sec:signal}). Run time considerations are discussed in
\secref{sec:runtime}, and \secref{sec:conclusion} presents final remarks and
future work.

\section{Related Work}
\label{sec:related}

While there is a large gamma of previous work for pupil detection, most methods
are not suitable for \emph{out-of-the-lab} scenarios.
For an extensive appraisal of state-of-the-art pupil detection methods, we refer
the reader to the works by \cite{fuhl2016pupil} and ~\cite{tonsen2016labelled}
for head-mounted eye trackers as well as ~\cite{fuhl2016evaluation} for remote
eye trackers.
In this work, we focus solely on methods that have been shown to be relatively
robust enough for deployment in pervasive scenarios, namely
\els~\citep{fuhl2016else}, \exc~\citep{fuhl2015excuse}, and
\swi~\citep{swirski2012robust}.

\begin{description}

\item[\els] consists of two approaches.
First, a Canny edge detector is applied, and the resulting edges are filtered
through morphological operations\footnote{\cite{fuhl2016else} also describe an
algorithmic approach to edge filtering producing similar results; however the
morphological approach is preferred because it requires less computing power.}.
Afterwards, ellipses are fit to the remaining edges, edges are removed based on
emprirically defined heuristics, and one ellipse is selected as pupil based on its
roundness and enclosed intensity value.
If this method fails to produce a pupil, a second approach that combines a mean
and a center surround filter to find a coarse pupil estimate is employed; an
area around this coarse estimate is then thresholded with an adaptive parameter,
and the center of mass of pixels below the threshold is returned as pupil center
estimate~\citep{fuhl2016else}.

\item[\exc] first analyzes the input images w.r.t. reflections based on
peaks in the intensity histogram.
If the image is determined to be reflection free, the image is thresholded with
an adaptive parameter, and a coarse pupil position is estimated through an
angular integral projection function~\cite{mohammed2012accurate}; this position
is then refined based on surrounding intensity values.
If a reflection is detected, a Canny edge detector is applied, and the resulting
edges are filtered with morphological operations; ellipses are fit to the
remaining edges, and the pupil is then selected as the ellipse with the darkest
enclosed intensity~\citep{fuhl2015excuse}.

\item[\swi] starts with a coarse positioning using Haar-like features. The
intensity histogram of an area around the coarse position is clustered using
\emph{k-means} clustering, followed by a modified \emph{RANSAC}-based
ellipse fit~\citep{swirski2012robust}.

\end{description}

From these algorithms, \els{} has shown a significantly better performance over
multiple data sets~\citep{fuhl2016pupil}.
Moreover, it is worth noticing that these algorithms employ multiple parameters
that were empirically defined, albeit there is usually no need to tune these
parameters.

It is worth dedicating part of this section to discuss machine-learning
approaches in contrast to the algorithmic ones, particularly convolutional
neural networks (CNN).
Similarly to other computer vision problems, from a solely pupil detection stand
point, deep CNNs will likely outperform human-crafted pupil detection approaches
given enough training data -- with incremental improvements appearing as more
data becomes available and finer network tuning.
Besides labeled data availability, which might be alleviated with
developments of unsupervised learning methods, there are other impediments to
the use of CNNs in pervasive scenarios -- i.e., in embedded systems.
For instance, computation time and power consumption. While these impediments
might be lessened with specialized hardware -- e.g.,
\emph{cuDNN}~\citep{chetlur2014cudnn}, \emph{Tensilica Vision
DSP}~\citep{efland2016high}, such hardware might not always be available or
incur prohibitive additional production costs.
Finally, CNN-based approaches might be an interesting solution from an
engineering point of view, but remain a \emph{black box} from the scientific
one.
To date, we are aware of two previous works that employ CNNs for pupil
detections:
1) \pup~\citep{fuhl2016pupilnet}, which aims at a computationally inexpensive
solution in the absence of hardware support, and
2) \vo~\citep{vera2017deconvolutional}, which consists of two very deep CNNs --
a coarse estimation stage (with 35 convolution plus 7 max-pooling layers for
encoding and 10 convolution plus 7 deconvolution layers for decoding), and a
fine estimation stage (with 14 convolution plus 5 max-pooling layers for
encoding and 7 convolution plus 5 deconvolution layers).


\section{\pure{}: The \underline{Pu}pil \underline{Re}constructor}
\label{sec:method}

\begin{figure}[h]
	\begin{center}
		\subfloat[][\label{a}]{
			\includegraphics[width=.18\columnwidth]{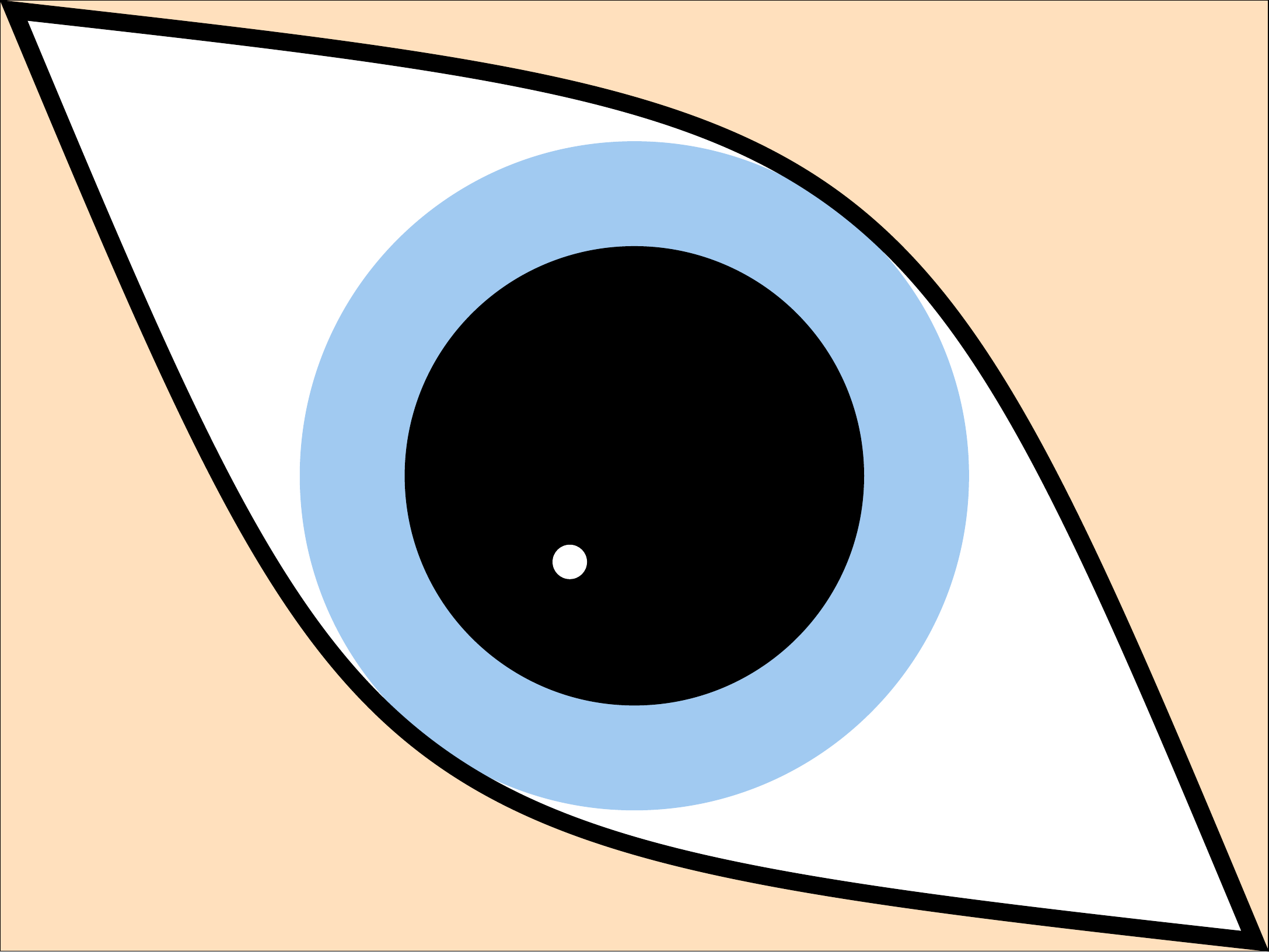}
			\label{fig:maxpupil}
		}
		\subfloat[][\label{b}]{
			\includegraphics[width=.18\columnwidth]{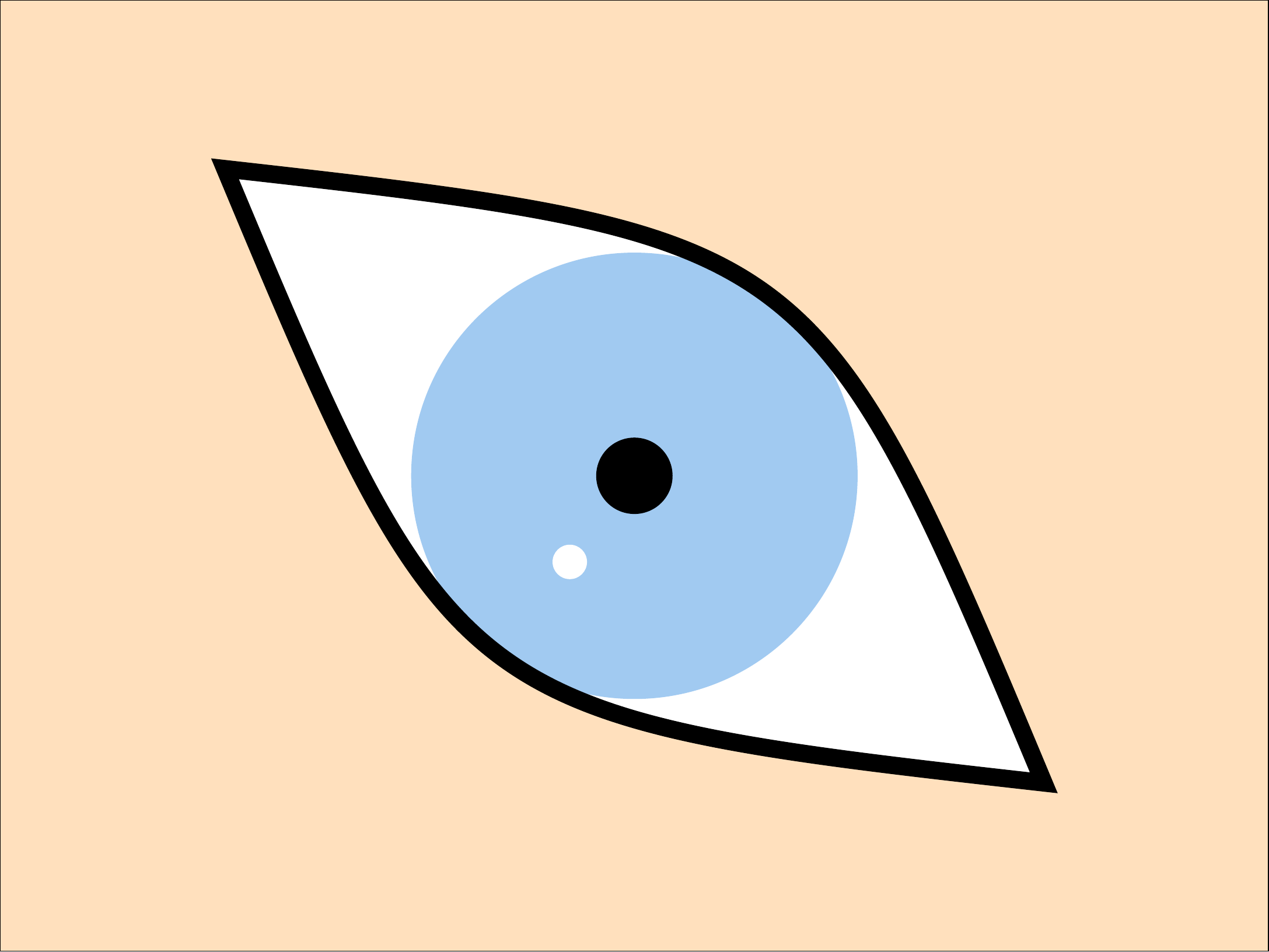}
			\label{fig:minpupil}
		}
		\subfloat[][\label{c}]{
			\includegraphics[width=.18\columnwidth]{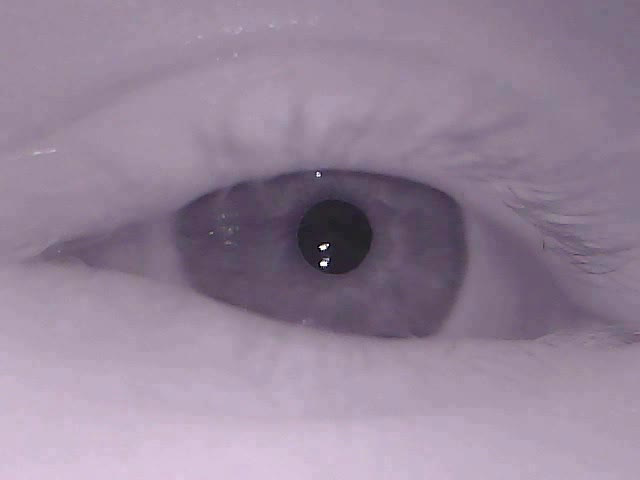}
			\label{fig:valid}
		}
		\subfloat[][\label{d}]{
			\includegraphics[width=.18\columnwidth]{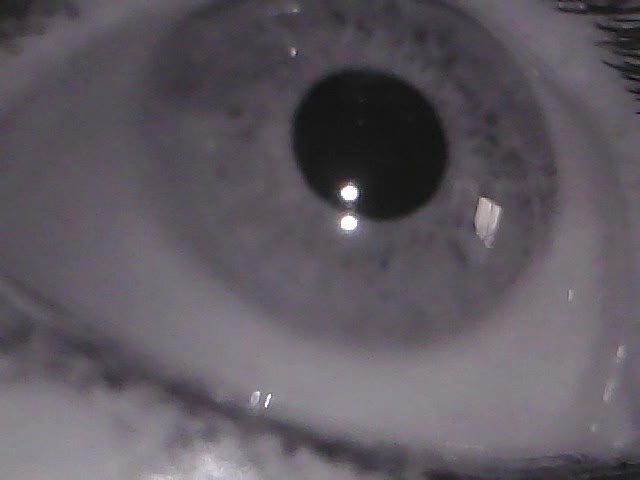}
			\label{fig:violation-valid}
		}
		\subfloat[][\label{e}]{
			\includegraphics[width=.18\columnwidth]{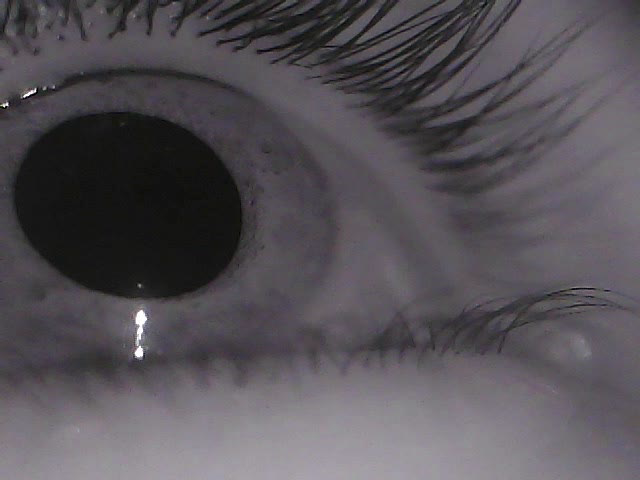}
			\label{fig:violation-invalid}
		}
		\caption{
			\pure{} assumptions visualized.  (a) illustrates the maximal intercanthal
			distance, yielding the \underline{max}imal \underline{p}upil
			\underline{d}iameter ($pd_{max}$), whereas
			(b) illustrates the lower bound -- i.e., \underline{min}imal
			\underline{p}upil \underline{d}iameter ($pd_{min}$).
			(c) shows realistic data that respects both assumptions. In
			contrast, the maximal intercanthal distance assumption is violated in (d) and (e).
			In the former, the pupil does not approach maximal dilation, and
			\pure{} is still able to detect the pupil.
			In the latter, the pupil is significantly dilated, and the resulting
			diameter exceeds $pd_{max}$; \pure{} does not detect such pupils.
		}
		\label{fig:inputex}
	\end{center}
\end{figure}

Similarly to related work, the proposed method was designed for
\emph{near-infrared}\footnote{This is the standard image format for head-mounted
eye trackers and can be compactly represented as a grayscale image.} eye images
acquired by head-mounted eye trackers.
Our method only makes two uncomplicated assumptions to constrain the valid pupil
dimension space without requiring empirically defined values: 1) the eye canthi
lay within the image, and 2) the eye canthi cover at least two-thirds of the
image diagonal.
It is worth noticing that these are \emph{soft} assumptions -- i.e., the
proposed method still operates satisfactorily if the assumptions are not
significantly violated.
\figref{fig:inputex} illustrates these concepts.
Furthermore, these assumptions are in accordance to eye tracker placement
typically suggested by eye tracker vendor's guidelines to capture the full
range of eye movements.

\pure{} works \emph{purely} based on edges, selecting curved edge segments that
are likely to be significant parts of the pupil outline.
These selected segments are then conditionally combined to construct further
candidates that may represent a reconstructed pupil outline.
An ellipse is fit to each candidate, and the candidate is evaluated
based on its ellipse aspect ratio, the angular spread of its edges relative to
the ellipse, and the ratio of ellipse outline points that support the hypothesis
of it being a pupil.
This evaluation yields a confidence measure for each candidate to be the pupil,
and the candidate with the highest confidence measure is then selected as pupil.
The remainder of this section describes the proposed method in detail.

\subsection{Preprocessing}
\label{sec:preprocessing}

Prior to processing, if required, the input image is downscaled to the
\underline{w}orking \underline{s}ize $S_w=( W_w \times H_w )$ through bilinear
interpolation, where $W_w$ and $H_w$ are the \underline{w}orking
\underline{w}idth and \underline{h}eight, respectively.
The original aspect ratio is respected during downscaling.
Afterwards, the resulting image is linearly normalized using a
\emph{Min-Max} approach.

\subsection{Edge Detection and Morphological Manipulation}
\label{sec:edgedetection}

\pure's first step is to perform edge detection using a Canny edge
operator~\citep{canny1986computational}.
The resulting edge image is then manipulated with a morphological approach to
thin and straighten edges as well as to break up orthogonal connections
following the procedure described by \cite{fuhl2016else}.
The result of this step is an image with unconnected and thinned edge segments.

\subsection{Edge Segment Selection}
\label{sec:selection}

Each edge segment is first approximated by a set of dominant points $D$
following the \emph{k-cosine} chain approximation method described by
\cite{teh1989detection}.
This approximation reduces the computational requirements for our approach and
typically results in a better ellipse fit in cases where a pupil segment has not
been properly separated from surrounding edges.
After approximation, multiple heuristics are applied to discard edge segments
that are not likely to be part of the pupil outline:

\begin{enumerate}

	\item Given the general conic equation $ ax^2+by^2+cxy+dx+ey+f=0$, at least
		five points are required to fit an ellipse in a least-squares sense.
		Therefore, we exclude segments in which $D$'s cardinality is smaller
		than five. This heuristic discards plain shapes such as small segments
		and substantially straight lines.

	\item Based on the assumptions highlighted in the beginning of this section,
		it is possible to establish the maximal and minimal distance between the
		lateral and medial \underline{e}ye \underline{c}anthus in pixels when
		frontally imaged as
		\begin{equation}
		ec_{max}=\sqrt{ {W_w}^2 + {H_w}^2 }
		\quad\text{and}\quad
		ec_{min}={{2}\over{3}}*ec_{max}.
		\end{equation}
		These estimates can then be used to infer rough values for the maximal
		(\figref{fig:maxpupil}) and minimal
		(\figref{fig:minpupil}) \underline{p}upil \underline{d}iameter bounds
		($pd_{max}$ and $pd_{min}$) based on the human physiology.
		We approximate the eye canthi distance through the palpebral fissure
		width as \SI{27.6}{\milli\meter}~\citep{kunjur2006anthropometric};
		similarly, the maximal and minimal pupil diameter are approximated as
		\SI{8}{\milli\meter} and \SI{2}{\milli\meter},
		respectively~\citep{spector1990clinical}.
		Therefore,
		\begin{equation}
		pd_{max}  \approx 0.29*ec_{max}
		\quad\text{and}\quad
		pd_{min}  \approx 0.07*ec_{min}.
		\end{equation}
		Note that whereas maximal values hold independent of camera rotation and
		translation w.r.t. the eye, minimal values might not hold due to
		perspective projection distortions and corneal refractions.
		Nonetheless, $pd_{min}$ already represents a minute part ($\approx
		4.8\%$) of the image diagonal, and we opted to retain this lower bound
		-- for reference, see \figref{fig:minpupil}.
		For each candidate, we approximate the segment's diameter by the largest
		gap between two of its points.
		Candidates with a diameter outside of the range $[pd_{min}, pd_{max}]$
		violate bounds and are thus discarded.

	\item To estimate a segment's curvature, first the minimum rectangle
		containing $D$ is calculated using the \emph{rotating calipers}
		method~\citep{toussaint1983solving}. The curvature is then estimated
		based on the ratio between this rectangle's smallest and largest
		sides.  The straighter the candidate is, the smaller the ratio.
		The cut-off threshold for this ratio is based on the ratio between the
		minor and major axes of an ellipsis with axes extremities inscribed in
		\SI{45}{\degree} of a circle, which evaluates to $R_{th} =
		(1-cos(\SI{22.5}{\degree}))/sin(\SI{22.5}{\degree}) \approx 0.2$.  This
		heuristic servers to discard relatively linear candidates.

	\item At this stage, an ellipse $E$ is fit to the points in $D$ following
		the least-squares method described in \citep{fitzgibbon1995buyer}. A segment is
		discarded if: I) $E$'s center lays outside of the image boundaries,
		which violates \pure's assumptions, or II) the ratio between $E$'s minor
		and major axes is smaller than $R_{th}$, which assumes that the camera
		pose relative to the eye can only distort the pupil round shape to a
		certain extend.

	\item Seldom, the ellipse fit will not produce a proper fit. We identify and
		discard most such cases inexpensivly if the mean point from $D$ does not lay
		within the polygon defined by the extremities of $E$'s axes.

\end{enumerate}

\subsection{Confidence Measure}
\label{sec:confidence}

For each remaining candidate, \pure{} takes into account three distinct metrics to
determine a confidence measure $\psi$ that the candidate is a pupil:

\begin{description}
	\item[Ellipse Aspect Ratio ($\rho$):] measures the roundness of $E$. This
		metric favors rounder ellipses (that typically result due to the eye
		camera placement w.r.t. the eye) and is evaluated as the ratio between
		$E$'s minor and major axis.

	\item[Angular Edge Spread ($\theta$):] measures the angular spread of the
		points in $D$ relative to $E$, assuming that the better distributed the
		edges are, the more likely it is that the edges originated from a
		clearly defined elliptical shape (i.e., a pupil's shape).
		This metric is roughly approximated as the ratio of $E$ centered quadrants
		that contain a point from $D$.

	\item[Ellipse Outline Contrast ($\gamma$):] measures the ratio of the $E$'s
		outline that supports the hypothesis of a darker region surrounded by a
		brighter region (i.e., a pupil's appearance).
		This metric is approximated by selecting $E$'s outline points with a
		stride of ten degrees.
		For each point, the linear equation passing through the point and the
		$E$'s center is calculated, which is used to define a line segment
		with length proportional to $E$'s minor axis and centered at the
		outline point.
		If the mean intensity of the inner segment is lower than the mean
		intensity of the outer one, the point supports the pupil-appearance
		hypothesis\footnote{If a bright pupil eye tracker is used, the inverse
		holds}.

\end{description}

If the candidate's ellipse outline is invalid -- i.e., violates \pure's size
assumptions or less than half of the outline contrast $\gamma$ supports the
candidate -- the confidence metric is set to zero.
Otherwise, the aforementioned metrics are averaged when determining the
resulting confidence. In other words,
\begin{equation}
	\psi =
	\begin{cases}
		0 & \mbox{if the outline is invalid;}\\
		{\rho + \theta + \gamma}\over{3} & \mbox{otherwise.}\\
	\end{cases}
\end{equation}
It is worth noticing that the range of all three metrics (and consequently
$\psi$) is [0,1].

\subsection{Conditional Segment Combination}
\label{sec:combination}

The segments that remain as candidates are combined pairwise to generate
additional candidates. This procedure attempts to reconstruct the pupil outline
based on nearby segment pairs since the pupil outline is often broken up due to
occlusions from, for example, reflections or eye lashes.
Let $D_1$ and $D_2$ be the set of dominant points for two segments and $S_1$ and
$S_2$ the set of points contained by the up-right squares bounding $D_1$ and
$D_2$, respectively.
The segments are combined if these bounding squares intersect but are not
fully contained into one another -- i.e., $S_1 \cap S_2 \ne \varnothing \ne S_1
\ne S_2$.
The resulting merged segment is then validated according to
\secref{sec:selection}, and its confidence measure evaluated according to
\secref{sec:confidence}.
Since this procedure is likely to produce candidates with high aspect ratio
$\rho$ and angular spread $\theta$ values, the new candidate is only added to
the candidate list if its outline contrast $\gamma$ improves on the $\gamma$
from the original segments.

After conditional combination, the candidate with highest confidence $\psi$ is
selected as the initial pupil.
Note that the inner intensities relative to other candidates do not contribute
to the pupil selection.
Thus, the iris might be selected since it exhibits properties similar
to the pupil -- e.g., roundness, inner-outer contrast, and size range.
For this reason, the inside of the initial pupil is searched for a roughly
cocentered candidate with adequate size and strong inner-outer outline contrast.
This is achieved through a circular search area centered at the center of the
initial pupil with radius equal to the initial semi-major axis -- i.e.,
representing a circular iris.
Candidates 1) lying inside this area, 2) with major axis smaller than the search
radius, and 3) with atleast three thirds of the outline contrast ($\gamma$)
valid are collected.
The collected candidate with highest confidence is then choosen as new
pupil estimate.
If no candidate is collected in this procedure, the initial pupil remains as the
pupil estimate.
As output, \pure{} returns not only a pupil center, but also its outline and a
confidence metric.

\section{Experimental Evaluation}
\label{sec:eval}

As previously mentioned, we evaluate \pure{} only against robust
state-of-the-art pupil detection methods, namely \els{}\footnote{With
morphological split and validity threshold.}, \exc{}, and \swi{}.
All algorithms were evaluated using their open-source \cpp{} implementations;
default parameters were employed unless specified otherwise.
For \exc{}, the input images were downscaled to \emph{240p} (i.e.,
\SI[product-units=single]{320 x 240}{\px}) as there is evidence that this is a
favorable input size detection-rate-wise~\citep{tonsen2016labelled}.
Similarly, the working size for \pure{} ($S_{w}$, \secref{sec:preprocessing})
was set to \emph{240p} as well to keep run time compatible with state-of-the-art
head-mounted eye trackers (see \secref{sec:runtime}).
\els{} provides an embedded downscaling and border cropping mechanism,
effectively operating with a resolution of \SI[product-units=single]{346 x
260}{\px}.
Notice that whenever the input images are downscaled, the results must be
upscaled to be compared with the ground truth.
No preprocessing downscaling was performed for \swi{} since evidence suggests it
degrades performance for this method~\citep{tonsen2016labelled}.
Additionally, we juxtapose our results to the ones from
\pup~\citep{fuhl2016pupilnet} and \vo~\citep{vera2017deconvolutional} whenever
possible.

In this work, we use the term \textbf{use case} to refer to each individual
eye video. For instance, the \lpw{} data set contains 22 subjects with three
recordings per subject in distinct conditions (e.g., indoors, outdoors),
resulting in 66 distinct \emph{use cases}.
Furthermore, we often compare \pure{} with the \textbf{rival}, meaning the
best performant from the other algorithms for the metric in question.
For instance, for the aggregated detection rate, \els{} performs better than
\exc{} and \swi{} and is, therefore, the \emph{rival}.

\subsection{Pupil Detection Rate}
\label{sec:detectionrate}

\begin{figure}[h]
	\centering
	\includegraphics[width=0.3\columnwidth]{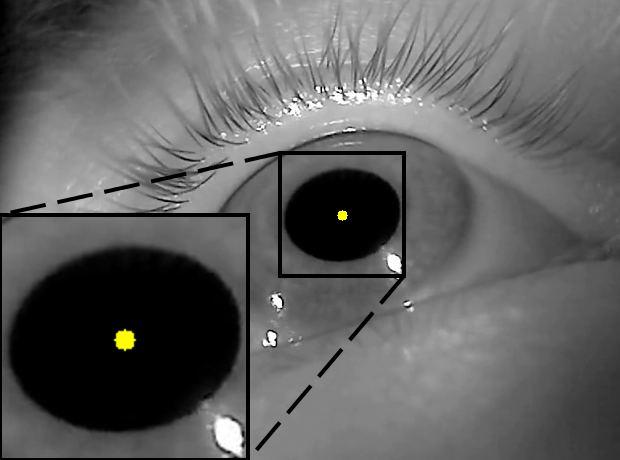}
	\hfill
	\includegraphics[width=0.3\columnwidth]{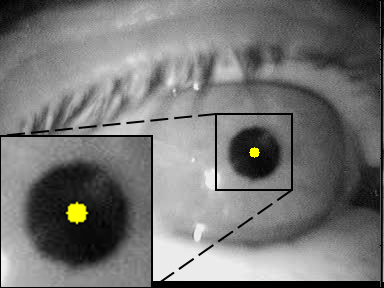}
	\hfill
	\includegraphics[width=0.3\columnwidth]{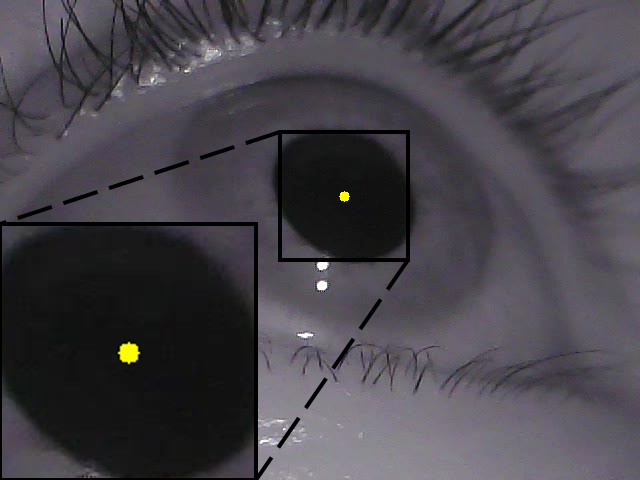}
	\caption{
		Five pixels validity range (in yellow) around the ground-truth pupil
		center for the pupil estimate to be considered correct and, thus, the
		pupil detected.
		Reference range relative to the data from the \swi{} (left),
		\exc/\els/\pup{} (center), and \lpw{} (right) data sets.
	}~\label{fig:validrange}
\end{figure}

A pupil is considered detected if the algorithm's pupil center estimate lies
within a radius of $n$ pixels from the ground-truth pupil center.
Similar to previous work, we focus on an error up to five pixels to account for
small deviations in the ground-truth labeling process -- e.g., human inaccuracy
\citep{fuhl2015excuse,fuhl2016else,tonsen2016labelled,vera2017deconvolutional}.
This error magnitude is illustrated in \figref{fig:validrange}.
For this evaluation, we employed five data sets totaling 266,786 realistic and
challenging images acquired with three distinct head-mounted eye tracking
devices, namely, the
\swi{} \citep{swirski2012robust},
\exc{} \citep{fuhl2015excuse},
\els{} \citep{fuhl2016else},
\lpw{} \citep{tonsen2016labelled},
and \pup{} \citep{fuhl2016pupilnet} data sets.
In total, these data sets encompass 99 distinct \emph{use cases}.
It is worth noticing that we corrected a disparity of one frame in the
ground truth for five \emph{use cases} of the \els{} data set and for the whole
\pup{} data set, which increased the detection rate of all algorithms (by
$\approx3.5\%$ on average).

\begin{figure}[h]
\centering
	\includegraphics[height=0.165\textheight]{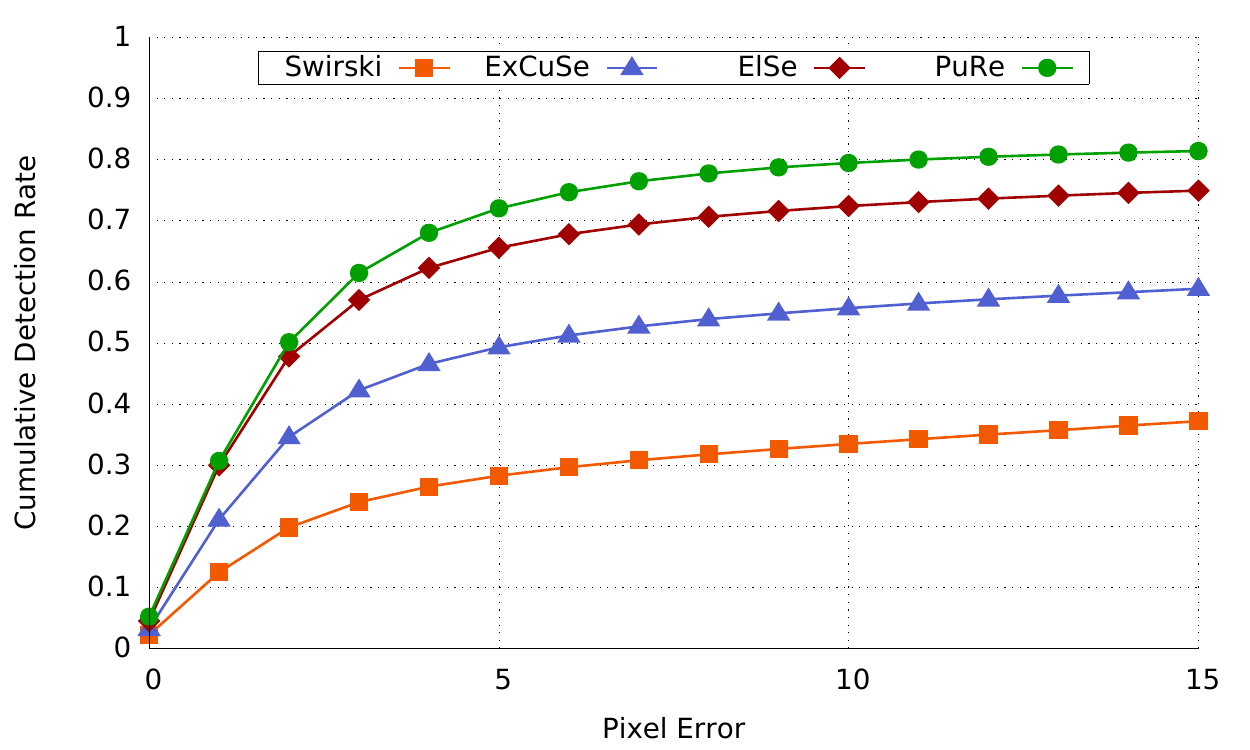}
	\hfill
	\includegraphics[height=0.165\textheight,trim={0 0 9cm 0},clip]{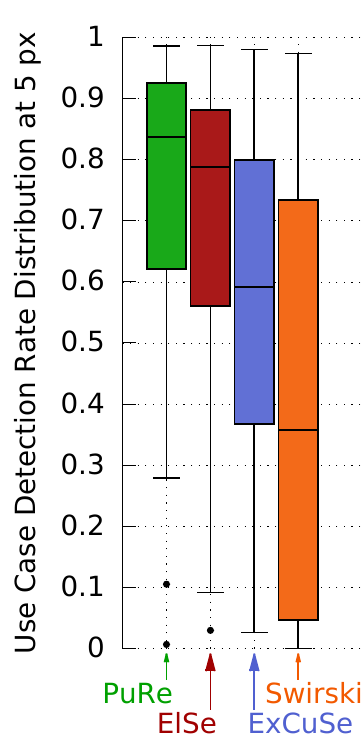}
	\caption{
		On the left, the cumulative detection rate for the aggregated 266,786
		images from all data sets. On the right, the distribution of the
		detection rate per \emph{use case} as a \emph{Tukey
		boxplot}~\protect\citep{frigge1989some}.
	}~\label{fig:cdr}
\end{figure}

\figref{fig:cdr} shows the cumulative detection rate per pixel error of the
evaluated algorithms for the aggregated 266,786 images as well as the detection
rate distribution per \emph{use case} at five pixels.
As can be seen, \pure{} outperforms all algorithmic competitors for all pixel
errors.
In particular, \pure{} achieved a detection rate of $72.02\%$ at the five pixel
error mark, further advancing the state-of-the-art detection rate by a
significant margin of 6.46 percentage points when compared to the \emph{rival}.
Moreover, the proposed method estimated the pupil center correctly $80\%$ of the
time for the majority of \emph{use cases}, attesting for \pure's comprehensive
applicability in realistic scenarios.

\begin{figure}[h]
\centering
	\includegraphics[width=\columnwidth]{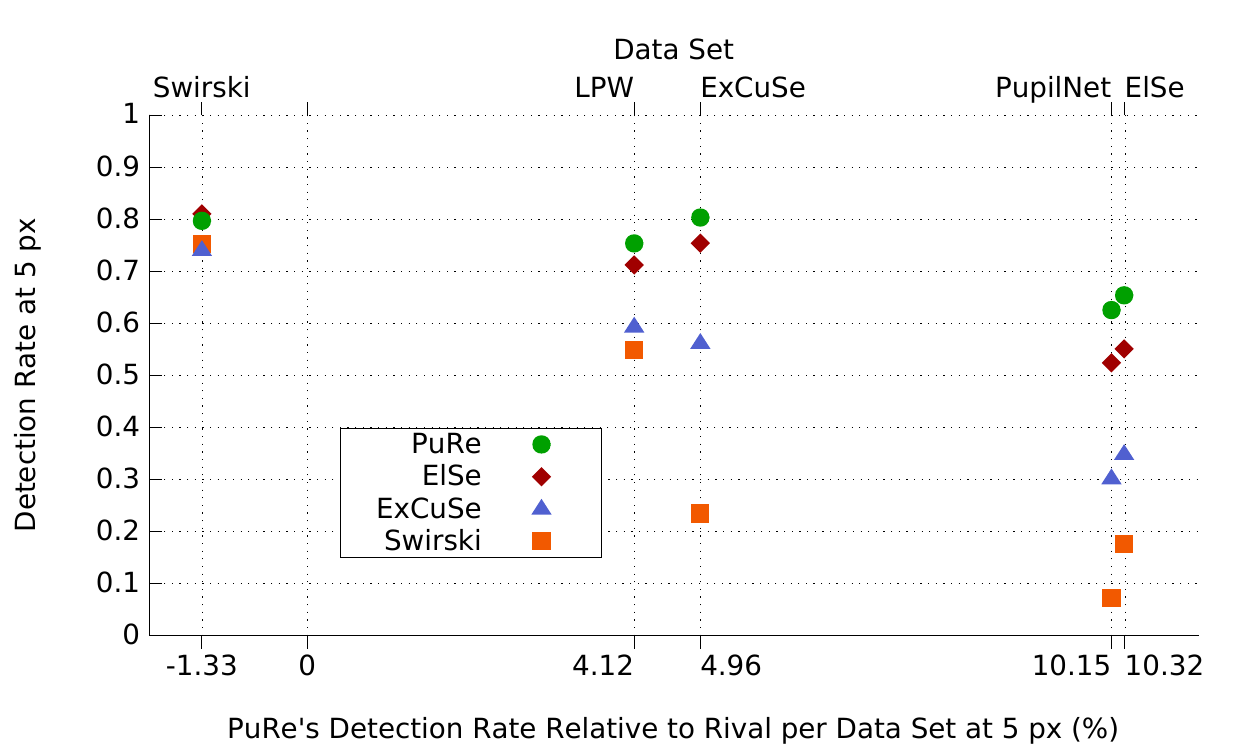}
	\caption{
		Detection rate per data set plotted against \pure's performance relative
		to the \emph{rival}. The lower the points, the harder the data set; the
		further right the points, the larger \pure's performance w.r.t. the
		\emph{rival} is. Notice that as the data sets become more difficult
		(i.e., the detection rate decreases for all algorithms), the gap
		between \pure{} and the other algorithms increases.
	}~\label{fig:drpd}
\end{figure}

\begin{figure}[htpb]
\centering
	\includegraphics[angle=-90,width=\columnwidth]{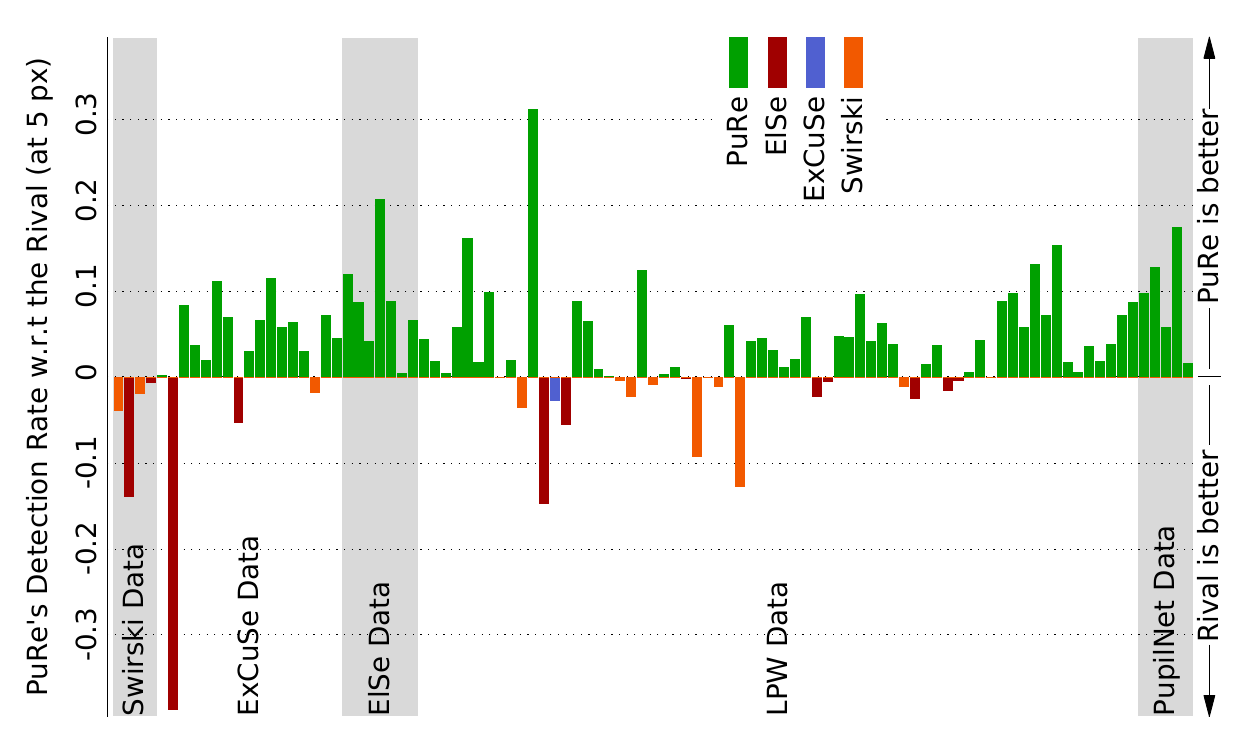}
	\caption{
		\pure's performance relative to the rival for each \emph{use case}.
		\pure{} is the best algorithm in $71.72\%$ of cases, \els{} in $14.14\%$,
		\swi{} in $12.12\%$, and \exc{} in $1.01\%$.
	}~\label{fig:reldr}
\end{figure}

\begin{figure}[htpb]
\centering
	\includegraphics[width=0.19\columnwidth]{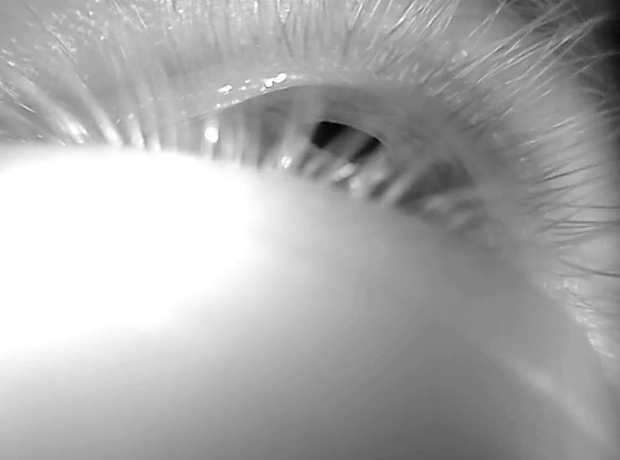}
	\includegraphics[width=0.19\columnwidth]{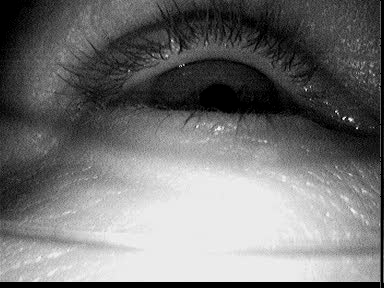}
	\includegraphics[width=0.19\columnwidth]{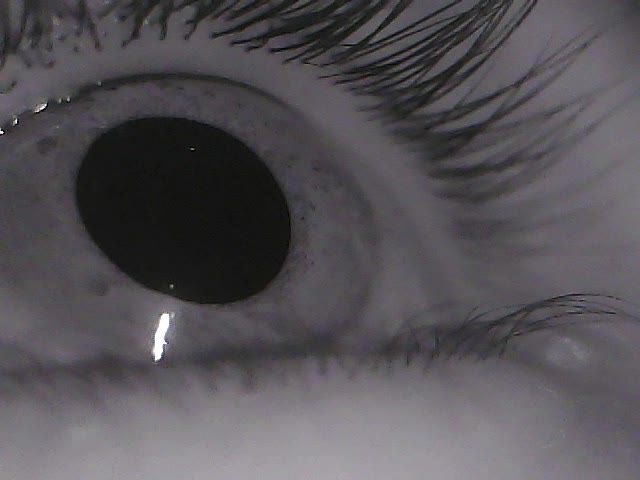}
	\includegraphics[width=0.19\columnwidth]{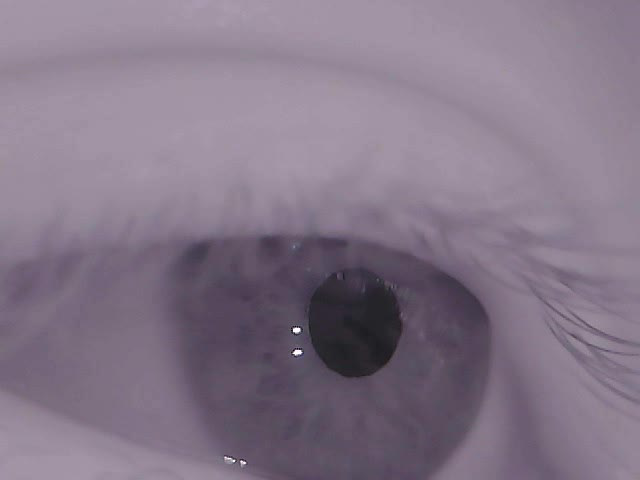}
	\includegraphics[width=0.19\columnwidth]{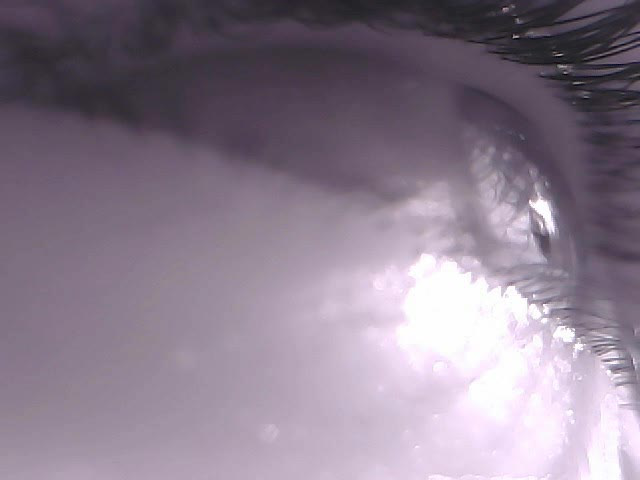}
	\\
	\includegraphics[width=0.19\columnwidth]{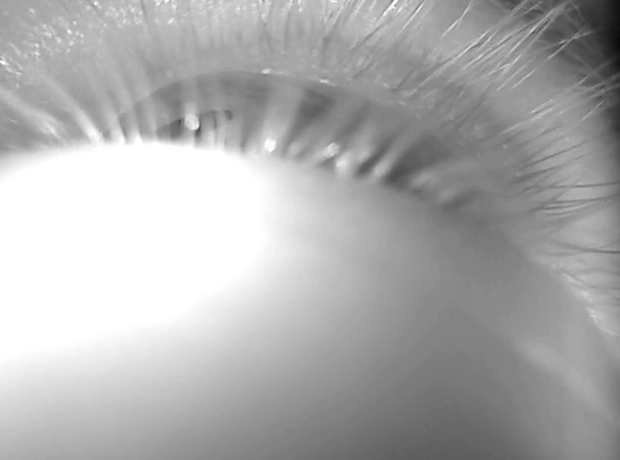}
	\includegraphics[width=0.19\columnwidth]{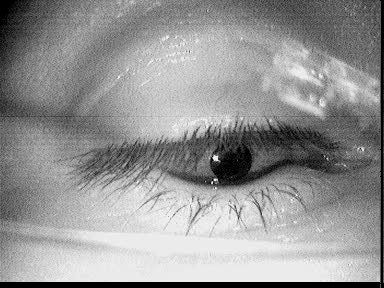}
	\includegraphics[width=0.19\columnwidth]{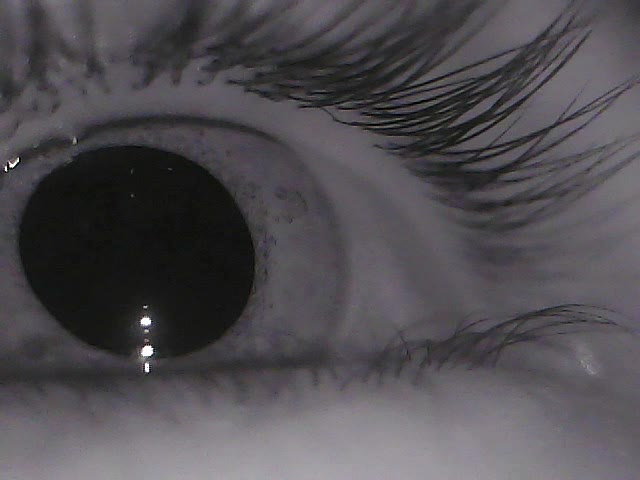}
	\includegraphics[width=0.19\columnwidth]{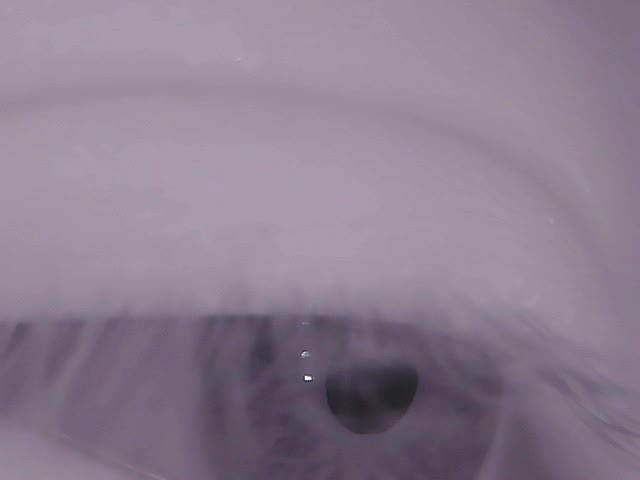}
	\includegraphics[width=0.19\columnwidth]{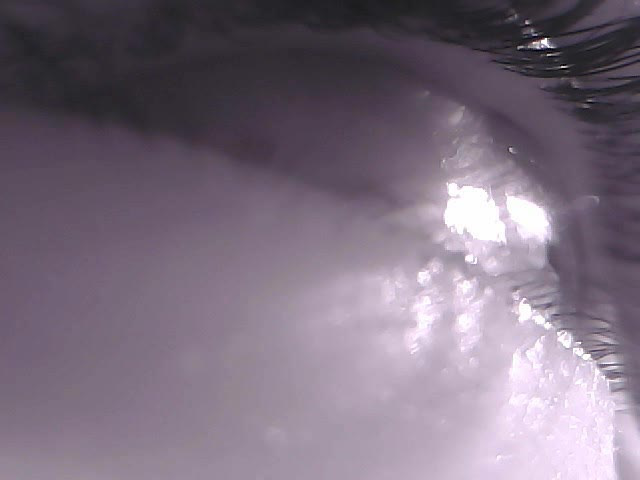}
	\\
	\includegraphics[width=0.19\columnwidth]{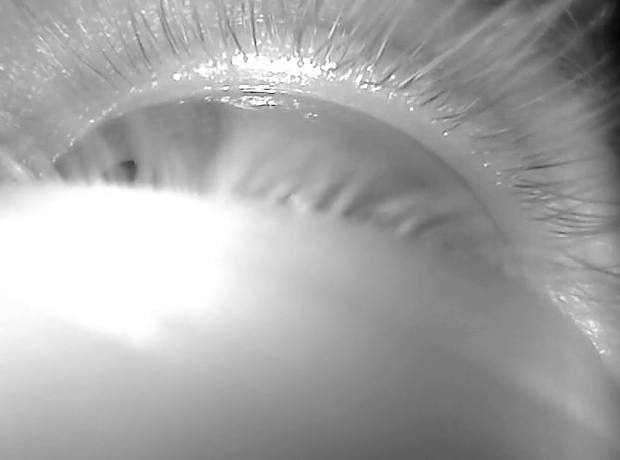}
	\includegraphics[width=0.19\columnwidth]{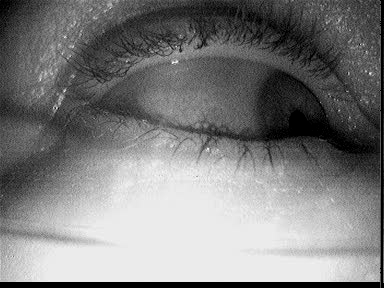}
	\includegraphics[width=0.19\columnwidth]{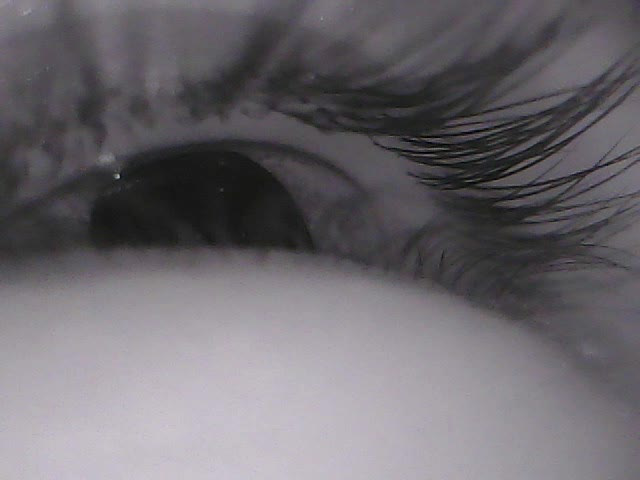}
	\includegraphics[width=0.19\columnwidth]{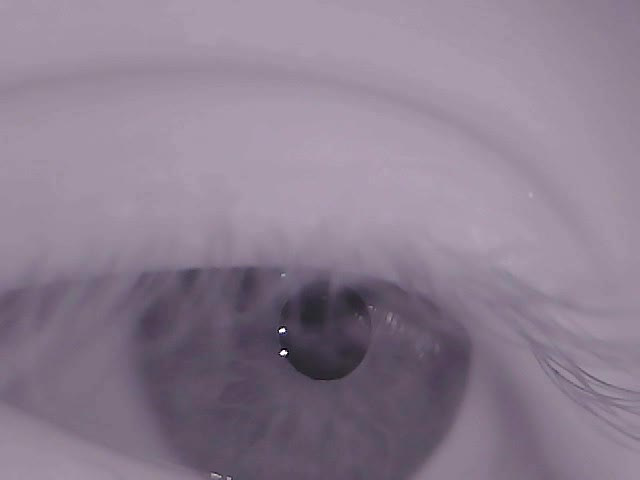}
	\includegraphics[width=0.19\columnwidth]{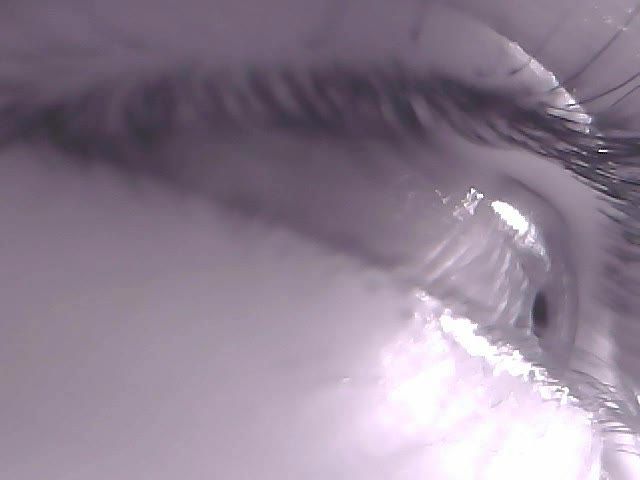}
	\caption{
		Representative frames for \emph{use cases} in which the \emph{rival}
		outperforms \pure. Each column contains frames from one \emph{use case}.
		From left to right: \ds{Swirski/p1-right}, \ds{ExCuSe/data-set-II},
		\ds{LPW/4/12}, \ds{LPW/9/17}, and \ds{LPW/10/11}.
	}~\label{fig:loses}
\end{figure}

It is worth noticing that the aggregated detection rate does not account for
differences in data set sizes.
As a consequence, this metric is dominated by the \exc{} and
\lpw{} datasets, which together represent $63.44\%$ of the data.
Since these two data sets are \emph{not the most challenging ones}\footnote{As
evidenced by higher detection rates for all algorithms in \figref{fig:drpd}}, the
algorithms tend to perform better on them, and differences between the
algorithms are less pronounced.
Inspecting the detection rates per data set in \figref{fig:drpd} gives a better
overview of the real differences between the algorithms and data sets, revealing that
\pure{} improves the detection rate by more than 10 percentage points w.r.t. the
\emph{rival} for the \emph{most challenging data sets} (i.e., \els{} and \pup).
To allow for a more fine-grained appreciation of the method's performance
relative to the other algorithms, \figref{fig:reldr} presents \pure's detection
rate at five pixels relative to the \emph{rival} for each \emph{use
case}. In $71.72\%$ of all \emph{use cases}, \pure{} outperformed all contenders.
In particular, for the two most challenging data sets, \pure{} surpassed the
competition in $100\%$ of the \emph{use cases}.
In contrast, the rivals noticeably outperformed \pure{} in five \emph{use cases}:
\ds{Swirski/p1-right}, \ds{ExCuSe/data-set-II}, \ds{LPW/4/12}, \ds{LPW/9/17},
and \ds{LPW/10/11}, from which representative frames are shown in
\figref{fig:loses}.
These five \emph{use cases} also highlight some of \pure's imperfections.
For instance, \ds{Swirski/p1-right} and \ds{ExCuSe/data-set-II} have weak and
broken pupil edges due to inferior illumination and occlusions due to eye
lashes/corneal reflections; \els{} compensates this lack of edges with its
second step.
\ds{LPW/4/12} contains large pupils that violate \pure's assumptions; in fact,
relaxing the maximum pupil size by only ten percent increases \pure's detection
rate from $44.2\%$ to $65.5\%$ (or $+6.55\%$ w.r.t. the \emph{rival}).
\ds{LPW/9/17} often has parts of the pupil outline occluded by eye lashes and
reflections, whereas \ds{LPW/10/11} contains pupils in extremely off-axial
positions combined with occlusions caused by reflections. However, visually
inspecting the latter two \emph{use cases}, we did not find any particular
reason for \swi{} to outperform the other algorithms.

Regarding CNN-based approaches\footnote{These approaches used the uncorrected \els{}
and \pup{} data sets, which might slightly affect the detection rate.}:
1) For \pup{}, \cite{fuhl2016pupilnet} report a detection rate of $65.88\%$ at the
five pixel error range when trained in half of the data from the \exc{} and
\pup{} data sets and evaluated on the remaining data. In contrast, \pure{}
reached $71.11\%$ on all images from these data sets -- i.e., $+5.23\%$.
2) For \vo{}, \cite{vera2017deconvolutional} report an unweighted\footnote{Averaged
over the \emph{use cases}.} detection rate of $82.17\%$ at the five pixel error
range averaged over a \emph{leave-one-out} cross validation in the \exc{} and
\els{} data sets. In contrast, \pure{} reached $76.71\%$ on all images from
these data sets -- i.e., $-5.46\%$.
Nevertheless, these results indicate that \pure{} is able to compete with
state-of-the-art CNN-based approaches while requiring only a small fraction of
CNN computational requirements. In fact, \pure{} outperformed \vo{} for $37.5\%$
of use cases.
Furthermore, it is worth noticing that the training data is relatively similar
to the evaluation data (same eye tracker, similar conditions and positioning) in
both cases, which might bias the results in favor of the CNN approaches.

\subsection{Beyond Pupil Detection Rate: Improving Precision, and Specificity
Through the Confidence Measure}
\label{sec:eval-conf}

One aspect that is often overlooked when developing pupil detection algorithms
is the rate of incorrect pupil estimates returned by the algorithm.
For instance, the aforementioned CNN-based approaches \emph{always} return a
pupil estimate, regardless of one actually existing in the image.
Intuitively, one can relax pupil appearance contraints in order to increase the
detection rate, leading to an increase in the amount of incorrect pupils
returned.
However, these incorrect pupil estimates later appear as noise and can
significantly degrade gaze-estimation calibration~\citep{santini2017calibme},
automatic eye movement detection~\citep{pedrotti2011data}, glanced-area ratio
estimations~\citep{vrzakova2012hard}, eye model
construction~\citep{swirski2013fully}, or even lead to wrong medical
diagnosis~\citep{jansen2009confidence}.
Therefore, it is imperative to also analyse algorithms in terms of incorrect
detected pupils.
The pupil detection task can be formulated as a classification problem --
similar to the approach by \cite{bashir2006performance} for frame-based tracking
metrics -- such that:
\begin{description}
	\item[True Positive (\tp)] represents cases in which the algorithm and
		ground truth agree on the presence of a pupil. We further specialize
		this class into \underline{C}orrect \underline{T}rue
		\underline{P}ositive (\ctp) and \underline{I}ncorrect \underline{T}rue
		\underline{P}ositive (\itp) following the detection definition from
		\secref{sec:detectionrate}.
	\item[False Positive (\fp)] represents cases in which the algorithm finds
		a pupil although no pupil is annotated in the ground truth.
	\item[True Negative (\tn)] represents cases in which the algorithm and
		ground truth agree on the absence of a pupil.
	\item[False Negative (\fn)] represents cases in which the algorithm fails
		to find the pupil annotated in the ground truth.
\end{description}
Note that this is not a proper binary classification problem, and the
\emph{relevant} class is given only by \ctp.
Therefore, we redefine \emph{sensitivity} and \emph{precision} in terms of this
class as
\begin{equation}
	sensitivity={{\mathit{CTP}} \over {\mathit{TP+FN}} }
\end{equation}
and
\begin{equation}
	precision={{\mathit{CTP}} \over {\mathit{TP+FP}} }
\end{equation}
respectively, such that \emph{sensitivity} reflects the (correct) pupil
detection rate and \emph{precision} the rate of pupils that the algorithm found
that are correct. Thus, these metrics allows us to evaluate 1) the
trade-off between detection of correct and incorrect pupils, and 2) the
meaningfulness of \pure's confidence measure.

\begin{figure}[h]
\centering
	\includegraphics[width=0.19\columnwidth]{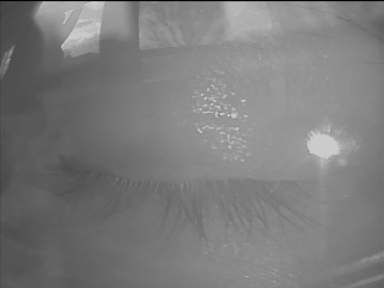}
	\includegraphics[width=0.19\columnwidth]{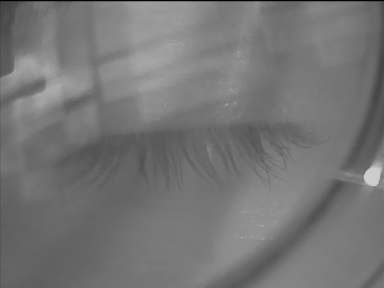}
	\includegraphics[width=0.19\columnwidth]{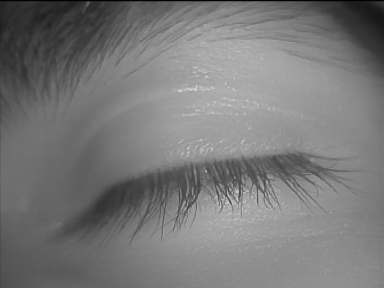}
	\includegraphics[width=0.19\columnwidth]{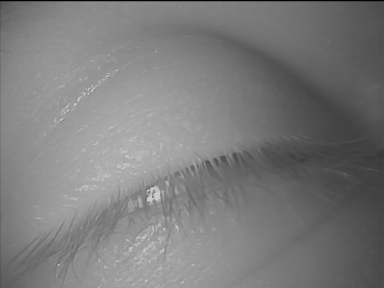}
	\includegraphics[width=0.19\columnwidth]{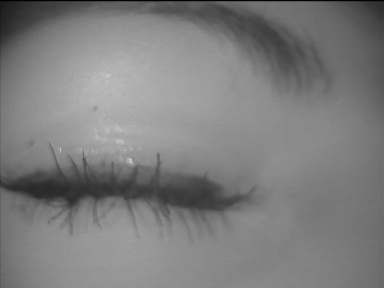}

	\includegraphics[width=0.19\columnwidth]{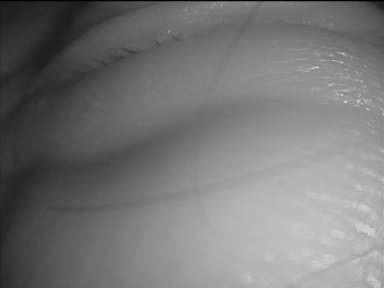}
	\includegraphics[width=0.19\columnwidth]{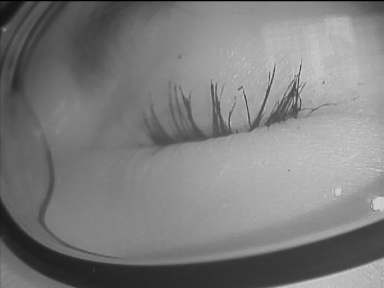}
	\includegraphics[width=0.19\columnwidth]{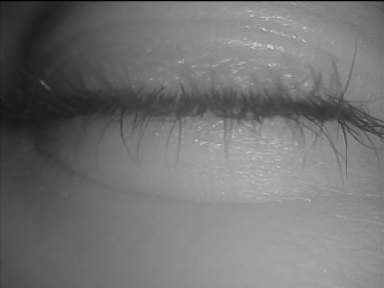}
	\includegraphics[width=0.19\columnwidth]{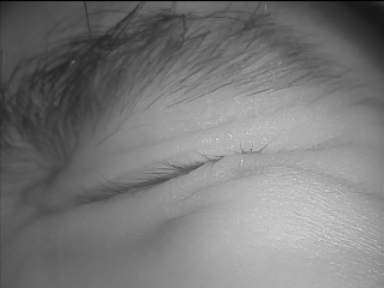}
	\includegraphics[width=0.19\columnwidth]{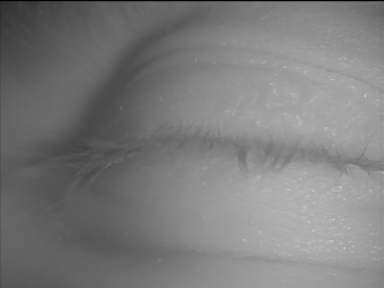}

	\caption{
		Samples from the \clo{} data set. First row shows samples from softly
		shut palpebrae, and the second one shows samples from strongly shut
		palpebrae.
	}~\label{fig:closedeyes}
\end{figure}

Unfortunately, the eye image corpus employed to evaluate pupil detection rates
(in \secref{sec:detectionrate}) do not include negative samples -- i.e., eye
images in which a pupil is not visible, such as during a blink.
Therefore, the capability of the algorithm to identify frames without a pupil as
such cannot be evaluated since \emph{specificity} $\big({{TN} \over {TN+FP}
}\big)$ remains undefined without negative samples.
To evaluate this aspect of the algorithms, we have recorded a new data set
(henceforth referred to as \clo{}) containing in its majority ($99.49\%$)
negative samples.
This data set consists of 83 \emph{use cases} and contains 49,790 images with a
resolution of \SI[product-units=single]{384 x 288}{\px}.
These images were collected from eleven subjects using a \emph{Dikablis
Professional} eye tracker~\citep{ergoneers2017} with varying illumination
conditions and camera positions.
A larger appearance variation was achieved by asking the subjects to perform
certain eye movement patterns\footnote{Although the eye is hidden underneath the
palpebrae, eye globe movement results in changes in the folds and light
reflections in the skin.} while their palpebrae remained shut in two
conditions: 1) with the palpebrae softly shut, and 2) with the palpebrae
strongly shut as to create additional skin folds.
In $\approx56\%$ of \emph{use cases}, participants wore glasses. Challenges in
the images include reflections, black eyewear frames, prominent eye lashes,
makeup, and skin folds, all of which can generate edge responses that the
algorithms might identify as parts of the pupil outline.
\figref{fig:closedeyes} shows representative images from the data set.

\begin{figure}[h]
\centering
	\includegraphics[width=\columnwidth]{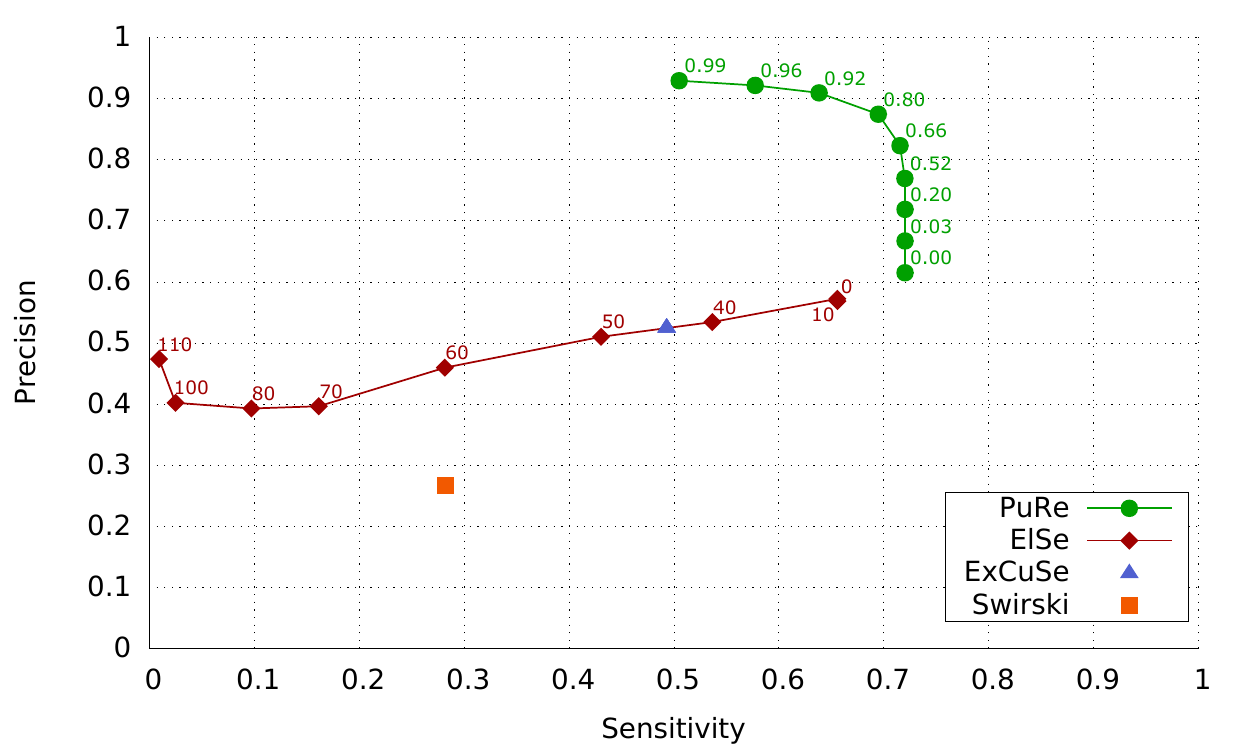}
	\includegraphics[width=\columnwidth]{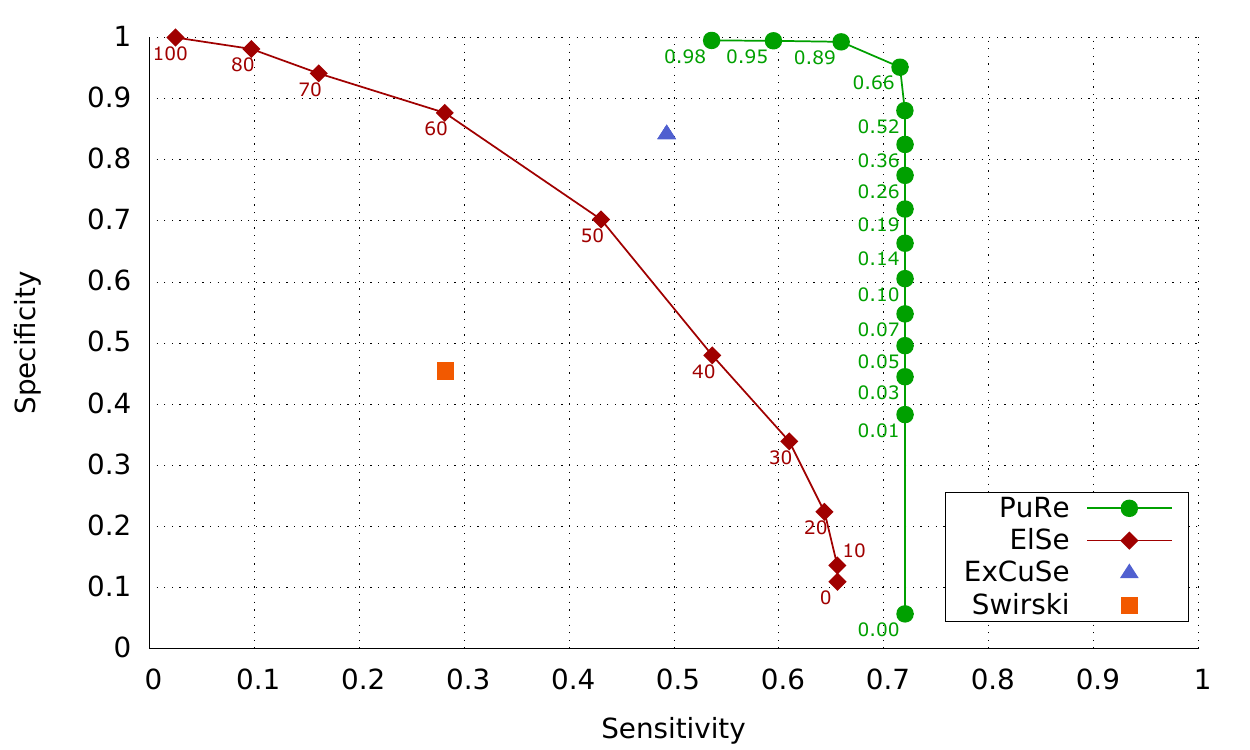}
	\caption{
		Trade-off between \emph{sensitivity} and \emph{precision} (top) as well
		as \emph{sensitivity} and \emph{specificity} (bottom) for different pupil
		validation thresholds for \pure{} and \els{}.
		Algorithms were evaluated over all images from the \swi{}, \exc{},
		\els{}, \lpw{}, \pup{}, and \clo{} data sets.
		For the sake of visibility, points are only plotted when there's a
		significant ($>0.05$) change in one of the metrics.
		The $Z_2$ score is maximized at thresholds 0.66 (for \pure) and 10 (for
		\els).
	}~\label{fig:confidence1}
\end{figure}

We evaluated the four aforementioned algorithms using all images from the
data sets from \secref{sec:detectionrate} and the \clo{} data set, totaling
316,576 images.
We assessed \pure's confidence measure using a threshold within [0:0.99] with
strides of 0.01 units.
A pupil estimate was considered correct only if its confidence measure was above
the threshold.
Similarly, \els{} offers a \emph{validity threshold} (default=10) to diminish
incorrect pupil rates, which we evaluated within the range [0:110] with strides
of 10 units.
\exc{} and \swi{} do not offer any incorrect pupil prevention mechanisms and,
therefore, result only in a single evaluation point.
The results from this evaluation are presented in \figref{fig:confidence1}.
As can be seen in this figure, \pure{} dominates over the other algorithms, and
\pure's confidence metric is remarkably meaningful, allowing to significantly
reduce incorrect pupil detections while preserving the correct pupil detection
rate and increasing identification of frames without pupils.
In fact, when compared to threshold 0, the threshold that maximizes the $F_2$
score (0.66) increased \emph{precision} and \emph{specificity} by $20.78\%$ and
$89.47\%$, respectively, whereas \emph{sensitivity} was decreased by a
negligible $0.49\%$.
In contrast, \els{} exhibited negligible ($<1\%$) changes for \emph{sensitivity}
and \emph{precision} when varying the threshold from 0 to 10, with a small gain
of $2.69\%$ in \emph{specificity}; subsequent increases in the threshold
increase \emph{specificity} at the cost of significantly deteriorating \els{}'s
performance for the other two metrics.
Compared to the \emph{rival} for each metric, \pure{}$_{th=0.66}$ improved
\emph{sensitivity}, \emph{precision}, and \emph{specificity} by $5.96$, $25.05$,
and $10.94$ percentage points, respectively.

\subsection{Pupil Signal Quality}
\label{sec:signal}

From the point of view of the image processing layer in the eye-tracking stack,
the (correct/incorrect) detection rates stand as a meaningful metric to measure
the quality of pupil detection algorithms.
However, the remaining layers (e.g., gaze estimation, eye movement
identification) often see the output of this layer as a discrete \emph{pupil
signal} (as a single-object tracking-by-detection), which these detection rates
do not fully describe.
For example, consider two pupil detection algorithms: \an{}, which detects
the pupil correctly every two frames, and \ap{}, which detects the pupil correctly
only through the first half of the data.
Based solely on the pupil detection rate (50\% in both cases), these algorithms
are identical.
Nonetheless, the former algorithm enables noisy\footnote{Note that the values
are not necessarily missing but might be incorrect pupil detections; thus
interpolation/smoothing might actually degrade the pupil signal even further.}
eye tracking throughout the whole data, whereas the latter enables noiseless eye
tracking during only the first half of the data.
Which algorithm is preferable is then application dependent, but a method to
assess these properties is required nonetheless.

Recent analyses of widely-used object tracking performance metrics have shown
that most existing metrics are strongly correlated and propose the use of only
two weakly-correlated metrics to measure tracker performance: \emph{accuracy}
and \emph{robustness}~\citep{vcehovin2016visual,kristan2016novel}.
Whereas in those works \emph{accuracy} was measured by average region overlaps, for
pupil detection data sets, only the pupil center is usually available.
Thus, we employ the center-error-based \emph{detection rate} as accuracy
measure.
As an indicator of \emph{robustness}, \cite{vcehovin2016visual} proposes the
\emph{failure rate} considering the tracking from a relibility engineering point
of view as a supervised system in which an operator reinitializes the tracker
whenever it fails.
For the pupil signal, we formulate this problem slightly different
since there is no operator reinitialization.
Instead, we evaluate the \emph{robustness} as the \emph{reliability}
\begin{equation}
	r = e^{ -\lambda t },
\end{equation}
where $\lambda = { {1} \over { \mtbf } } $ is the failure rate estimated through
the \underline{M}ean \underline{T}ime \underline{B}etween \underline{F}ailures
($\mtbf$) not accounting for \emph{repair time} -- i.e., periods of no/incorrect
pupil detection are considered as latent faults.
In this manner, the \emph{reliability} is a measure of the likelihood of the
algorithm to correctly detect the pupil for $t$ successive frames.
Furthermore, by measuring the \underline{M}ean \underline{T}ime \underline{T}o
\underline{R}epair ($\mttr$) -- i.e., the mean duration of periods in which the
correct pupil signal is not available -- we can achieve a similar metric in
terms of the likelihood for the algorithm to \emph{not} detect the pupil
correctly for $t$ successive frames.
Henceforth, we well define this metric as the \emph{insufficiency} ($i$), which
can be evaluated as
\begin{equation}
	i = e^{ -\kappa t },
\end{equation} where
$\kappa = { {1} \over { \mttr } } $. The smaller an algorithm's
\emph{insufficiency}, the more \emph{sufficient} it is.
It is worth noticing, that $r$ and $i$ are not true probabilities since the
events they measure are not likely to be independent nor uniformly distributed.
Consequently, these metrics only offer a qualitative and relative measure
between the algorithms.
Thus, we simplify their evaluation by fixing $t=1$.
As an illustration, let us return to our initial example considering a sequence of
$L$ frames:
\an{} yields $r_{A_1}=i_{A_1}=e^{-1/1}$, whereas \ap{} yields
$r_{A_2}=i_{A_2}=e^{-1/(0.5L)}$. Since $\forall \, L \, > \, 2 \, \implies \,
r_{A_1} \, < \, r_{A_2} \, \land \, i_{A_1} \, < \, i_{A_2}$, we can conclude that
\ap{} is more reliable but less sufficient w.r.t. \an{} for sequences longer
than two frames.
A quantitative conclusion is, however, not possible.

\begin{figure}[h]
\centering
	\includegraphics[width=\columnwidth]{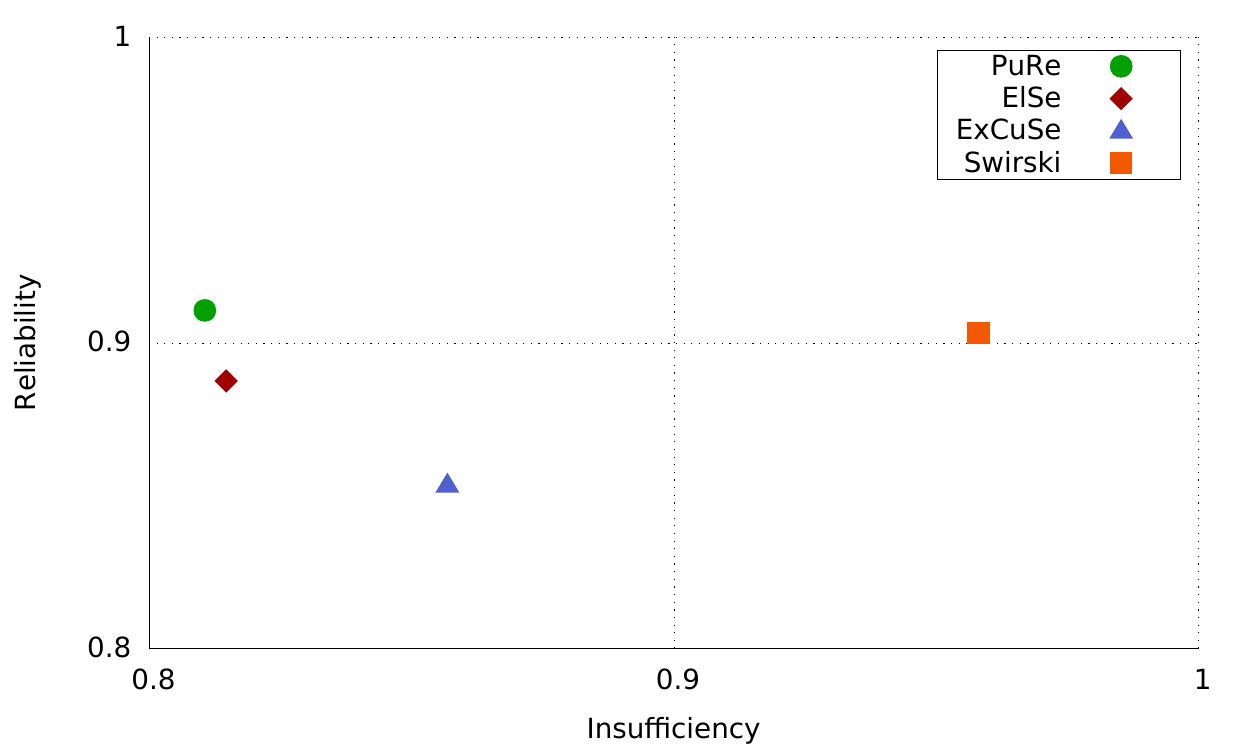}
	\caption{
		\emph{Reliability} (higher is better) and \emph{insufficiency} (lower is
		better) for all algorithms based on the sequence of all aggregated
		images from the \swi, \exc, \els, \lpw, and \pup{} data sets.
	}~\label{fig:dynamic}
\end{figure}

\begin{figure}[htpb]
\centering
	\includegraphics[angle=-90,width=.49\columnwidth]{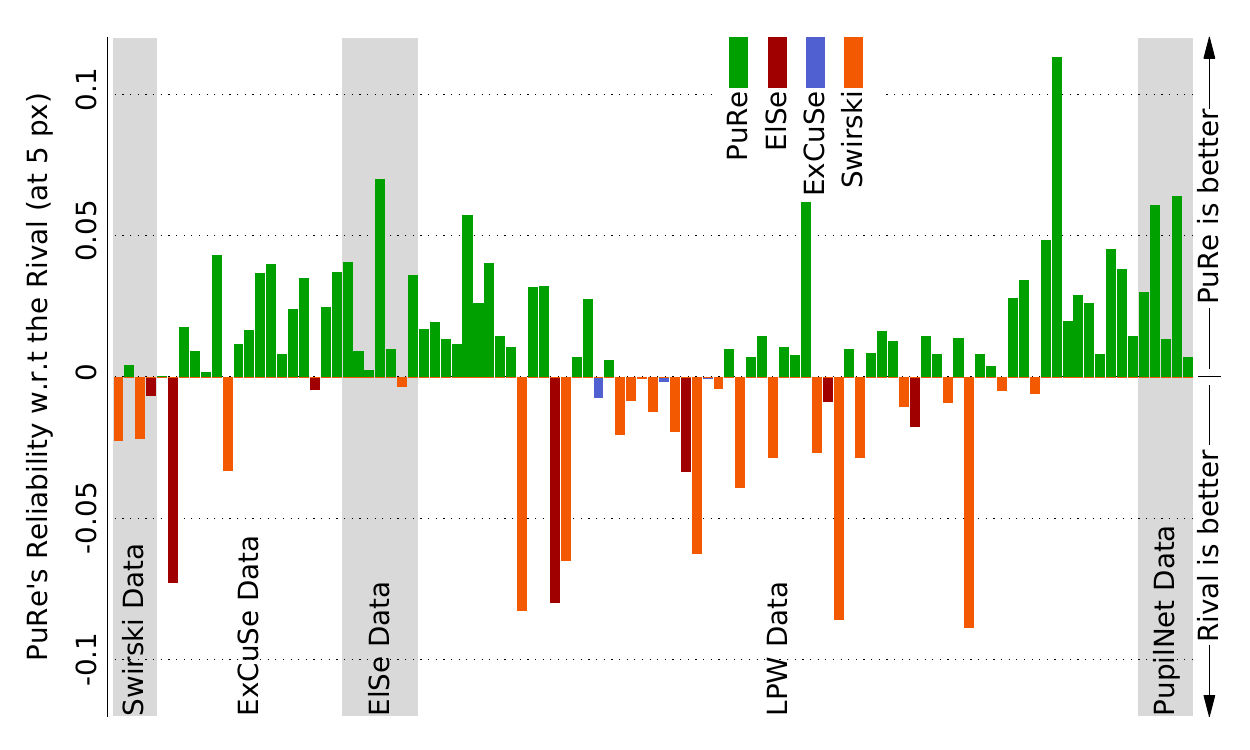}
	\includegraphics[angle=-90,width=.49\columnwidth]{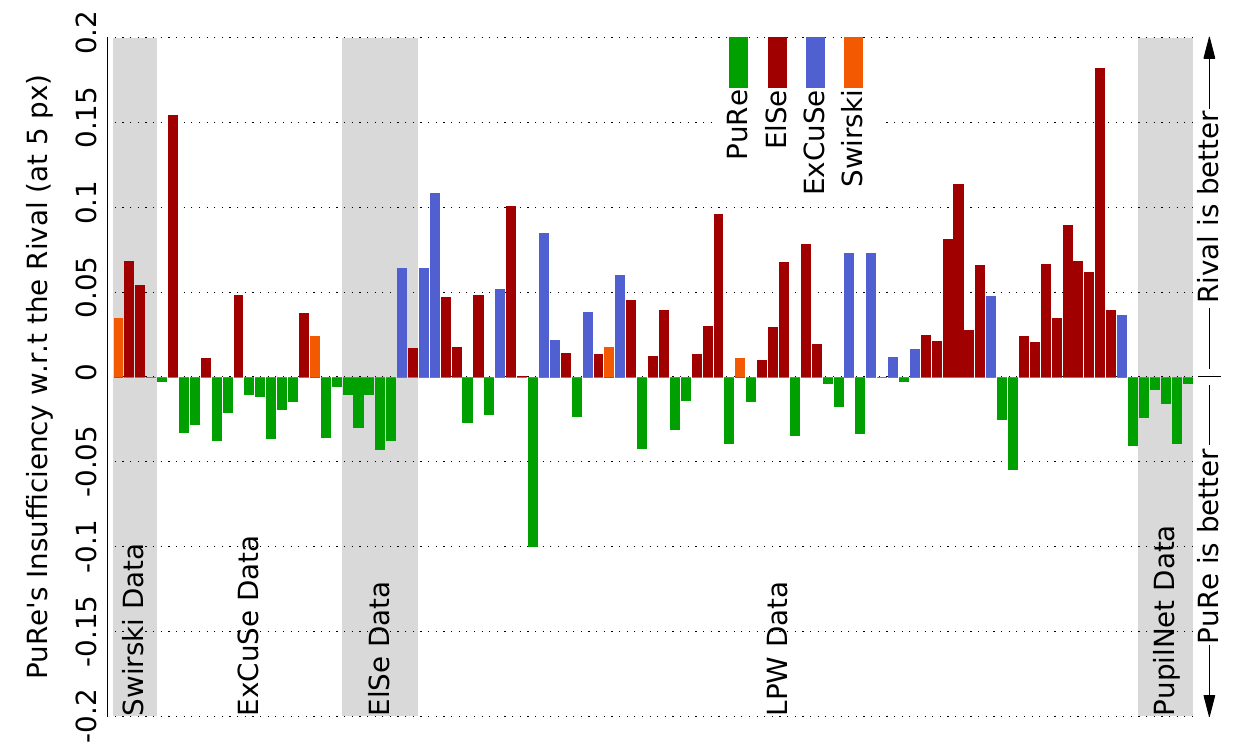}
	\caption{
		\pure's \emph{reliability} (left) and \emph{insufficiency} (right)
		relative to the rival for each \emph{use case}.
	}~\label{fig:pd-dynamic}
\end{figure}

We evaluated the four aforementioned algorithms in this regard using only the
data sets from \secref{sec:detectionrate}. The \clo{} data set was excluded
since it is not realistic from the temporal aspect  -- i.e., users are not
likely to have their eyes closed for extended periods of time.
Furhermore, it is worth noticing that each \emph{use case} from the \exc, \els,
and \pup{} data sets consists of images sampled throughout a video based on the
pupil detection failure of a commercial eye tracker; these \emph{use cases} can
be seen as videos with a low and inconstant sampling rate.
Results aggregated for all images are shown in \figref{fig:dynamic} and
indicate \pure{} as the most reliable and sufficient algorithm.
Curiously, the second most reliable algorithm was \swi, indicating that during
use cases in which it was able to detect the pupil, it produced a more stable
signal than \els{} and \exc{} -- although its detection rate is much lower
relative to the other algorithms for challenging scenarios.
This lower detection rate reflects on the \emph{insufficiency}, in which \swi{} is
the worst performer; \els{} places second, followed by \exc.
Furthermore, \figref{fig:pd-dynamic} details these results per \emph{use case}.
In this scenario, \pure{} was the most reliable algorithm in $66.67\%$ of the
\emph{use cases}, followed by \swi{} ($23.23\%$), \els{} ($7.07\%$), and \exc{}
($3.03\%$).
These results demonstrate that \pure{} is more reliable not only when taking
into account all images but also for the majority of \emph{use cases}.
This higher reliability also reflects on \pure's longest period of consecutive
correct pupil detections, which contained 859 frames (in \ds{LPW/21/12}). In
contrast, the longest sequence for the \emph{rival} was only 578 frames (\exc,
also in \ds{LPW/21/12}).
\els's longest period was of 386 frames in \ds{LPW/10/8}, for which
\pure{} managed 411 frames.
In terms of \emph{sufficiency} -- i.e., smaller \emph{insufficiency} -- \els{}
had a small lead with $41.41\%$ of \emph{use cases}, closely followed by \pure{}
($40.40\%$); \exc{} and \swi{} were far behind, winning $14.14\%$ and $4.04\%$
of use cases, respectively.
The advantage of \els{} here is likely due to its second pupil detection step,
which might return the correct pupil during periods of mostly incorrect
detections, fragmenting these periods into smaller ones.

\subsection{Run Time}
\label{sec:runtime}

\begin{figure}[h]
\centering
	\includegraphics[width=\columnwidth]{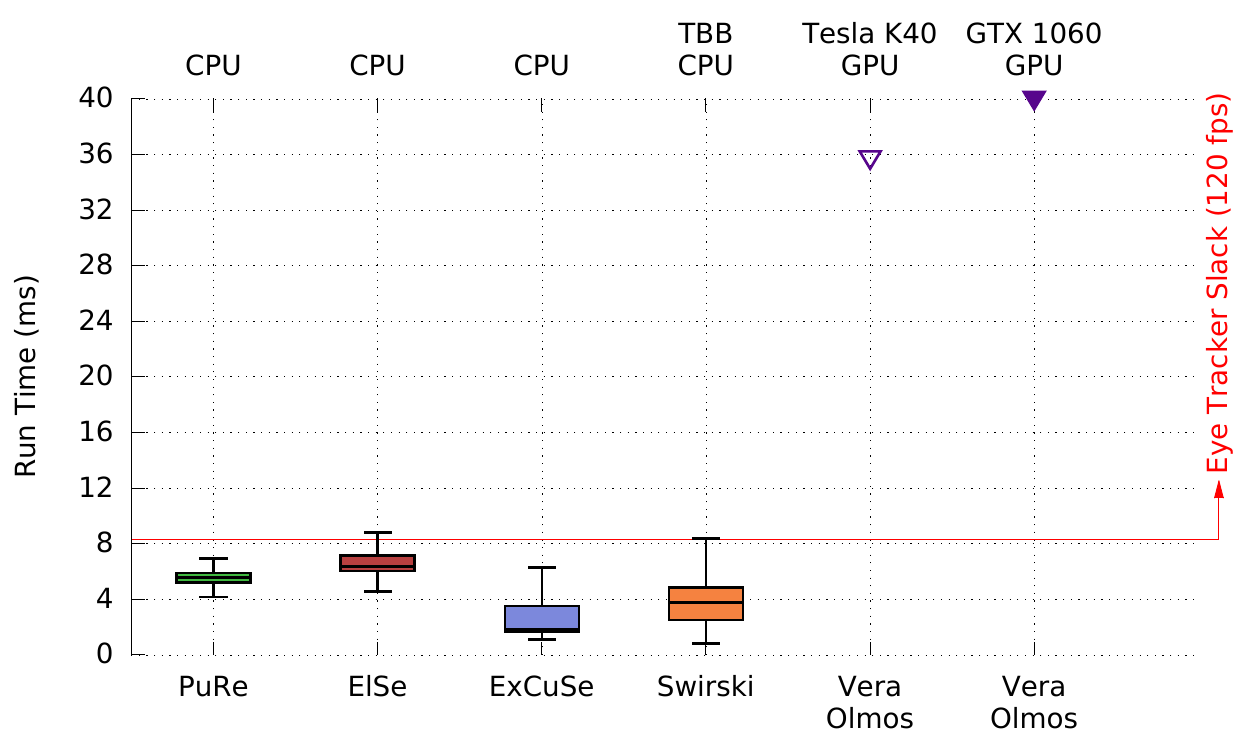}
	\caption{
		For \pure, \els, \exc, and \swi: Run time distribution across all images
		in the \swi{}, \exc{}, \els{}, \lpw{}, \pup{}, and \clo{} data sets.
		Note that these algorithms were evaluated on a CPU and only \swi{} was
		parallelized.
		For \vo: Run time as reported in~\cite{vera2017deconvolutional}, which
		were obtained with parallelized implementations using GPUs.
	}~\label{fig:runtime}
\end{figure}

The run time of pupil detection algorithms is of particular importance for
real-time usage -- e.g., for human-computer interaction. In this section, we
evaluate the temporal performance of the algorithms across all images from the
\swi{}, \exc{}, \els{}, \lpw{}, \pup{}, and \clo{} data sets.
Evaluation was performed on a Intel\textregistered{} Core\texttrademark{}
i5-4590 CPU @ 3.30GHz with 16GB RAM under Windows 8.1, which is similar to
systems employed by eye tracker vendors.
Results are shown in \figref{fig:runtime}. All algorithms exhibited competitive
performance in terms of run time, conforming with the slack required for
operation with state-of-the-art head-mounted eye trackers. For instance, the
\cite{pupillabs2017} eye tracker, which provides images at \SI{120}{\hertz} --
i.e., a slack of \SI{\approx8.33}{\milli\second}.
Henceforth, we will use the notation $\mu$ for the mean value and $\sigma$ for
the standard deviation.
Run time wise, \exc{} was the best performer ($\mu=2.51$, $\sigma=1.11$),
followed by \swi{} ($\mu=3.77$, $\sigma=1.77$), \pure{} ($\mu=5.56$,
$\sigma=0.6$), and \els{} ($\mu=6.59$, $\sigma=0.79$).
It is worth noticing that \els{} operates on slightly larger images
(\SI[product-units=single]{346 x 260}{\px}) w.r.t. \pure{} and \exc{}
(\SI[product-units=single]{320 x 240}{\px}).
Furthermore, \swi{} operates on the original image sizes, but its implementation
is parallelized using \emph{Intel Thread Building
Blocks}~\citep{pheatt2008intel}, whereas the other algorithms were not
parallelized.
In contrast to the algorithmic approaches, \cite{vera2017deconvolutional} report
run times for their CNN-based approach of $\approx\SI{36}{\milli\second}$ and
$\approx\SI{40}{\milli\second}$ running on a \emph{NVidia Tesla K40 GPU} and a
\emph{NVidia GTX 1060 GPU}, respectively.
It is worth noticing that these run times are still more than four times larger
than the slack required by modern eye trackers and almost one order of magnitude
larger than the algorithmic approaches running on a CPU.

\section{Final Remarks}
\label{sec:conclusion}

In this paper, we have proposed and evaluated \pure{}, a novel edge-based
algorithm for pupil detection, which significantly improves on the
state-of-the-art in terms of \emph{sensitivity}, \emph{precision}, and
\emph{specificity} by $5.96$, $25.05$, and $10.94$ percentage points,
respectively.
For the most challenging data sets, detection rate was improved by more than ten
percentage points.
\pure{} operates in real-time for modern eye trackers (at 120 \emph{fps}).
An additional contribution was made in the form of new metrics to evaluate
pupil detection algorithms.

Furthermore, results show that our single-method edge-based approach outperformed even
two-method approaches (e.g., \els{} and \exc).
However, there are clear (but uncommon) cases when an edge-based approach will
not suffice due to lack of edge-information in the image.
For instance, extremely blurred images, or if a significant part of the pupil outline
is occluded.
In such cases, \pure{} offers a meaningful confidence measure for the
detected pupil, which can be used to identify whenever \pure{} failed.
Following from our analysis in \secref{sec:eval-conf}, we recommend a threshold
of 0.66 for this confidence measure.
Thus, whenever \pure{} can not find a pupil, an alternative pupil detection
method can be employed -- e.g., \els's fast second step.
Nonetheless, care has to be taken not to compromise \emph{specificity} through
this second step.

\begin{figure}[h]
\centering
	\includegraphics[width=0.19\columnwidth]{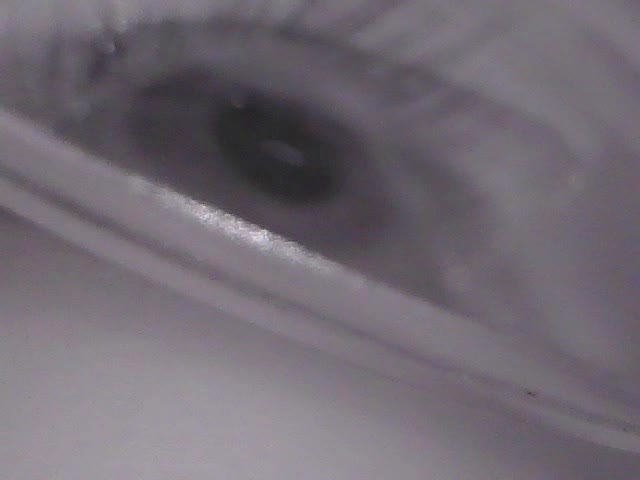}
	\includegraphics[width=0.19\columnwidth]{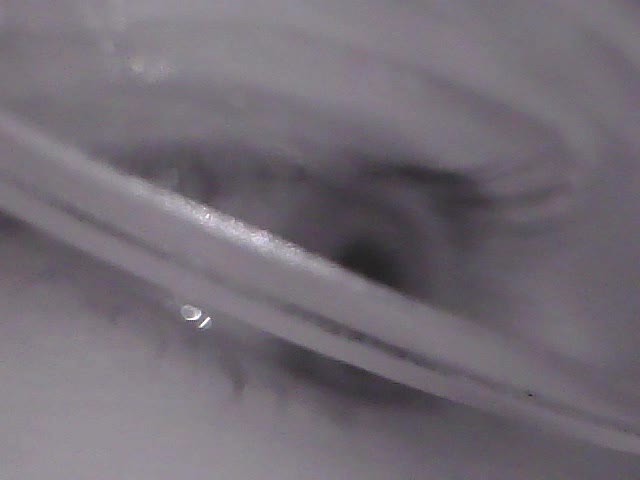}
	\includegraphics[width=0.19\columnwidth]{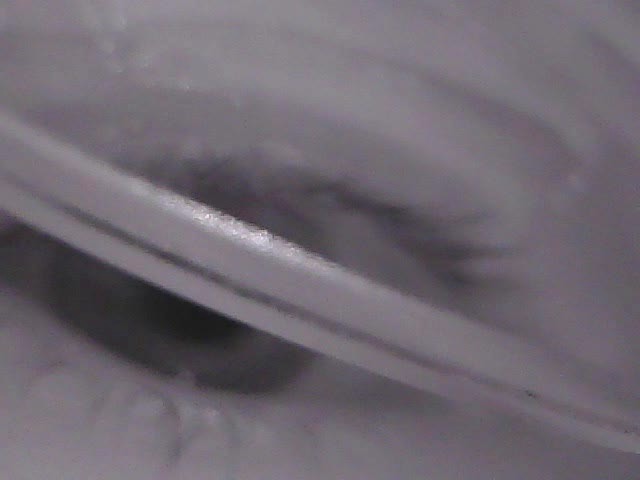}
	\includegraphics[width=0.19\columnwidth]{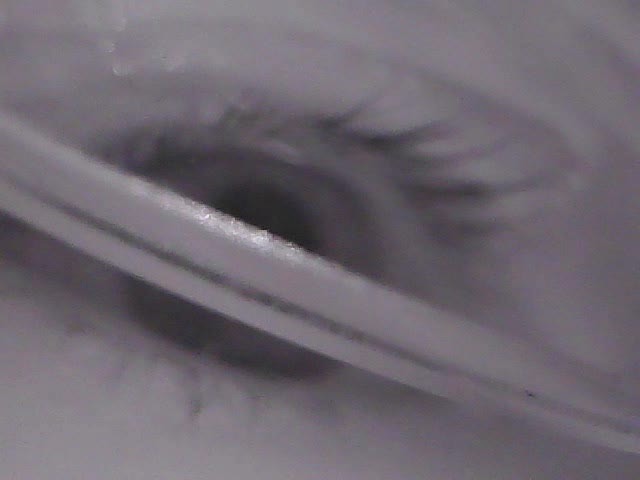}
	\includegraphics[width=0.19\columnwidth]{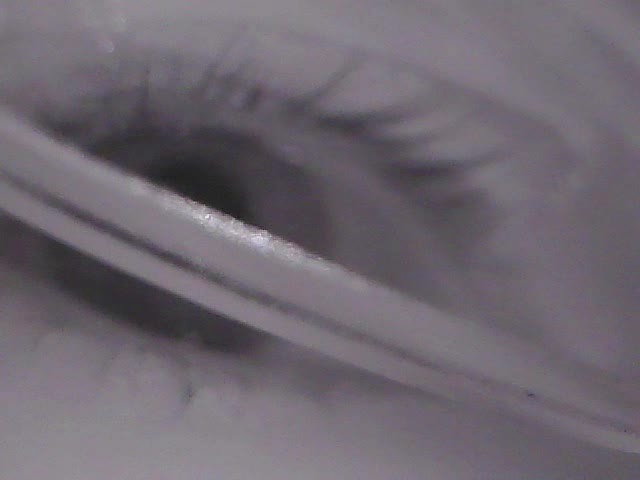}

	\includegraphics[width=0.19\columnwidth]{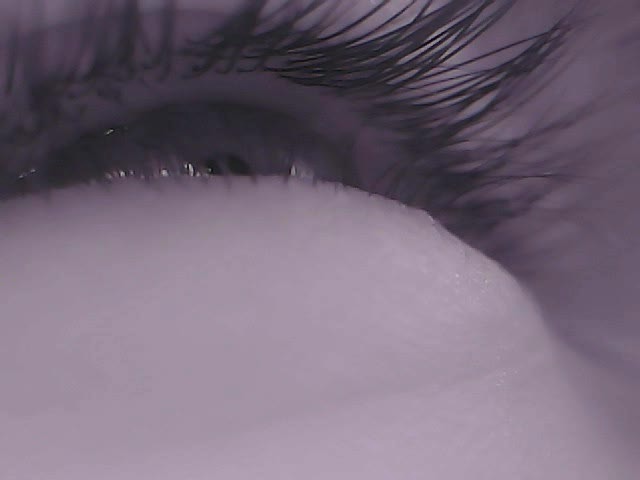}
	\includegraphics[width=0.19\columnwidth]{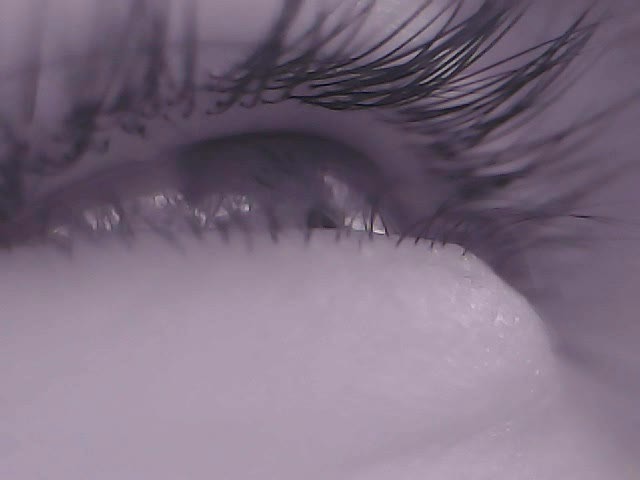}
	\includegraphics[width=0.19\columnwidth]{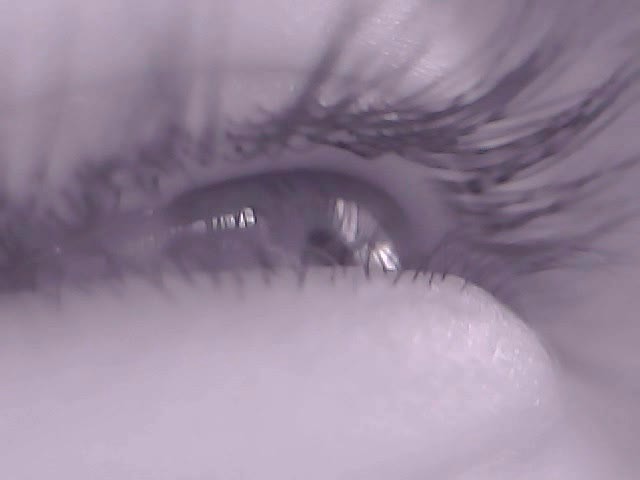}
	\includegraphics[width=0.19\columnwidth]{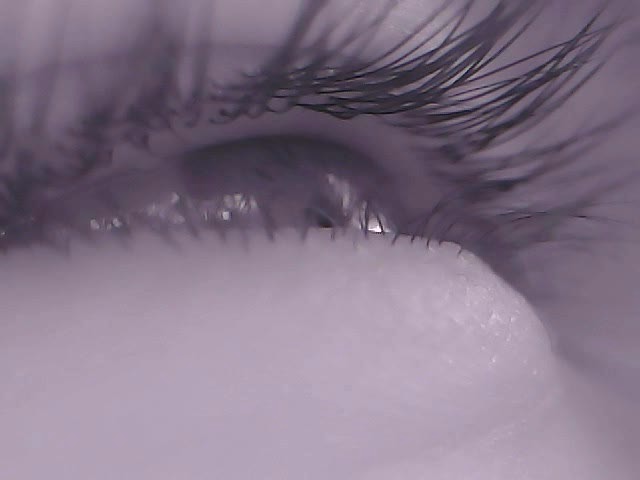}
	\includegraphics[width=0.19\columnwidth]{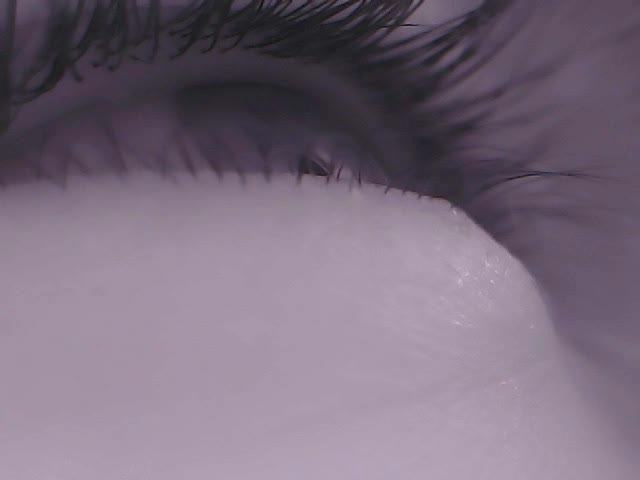}

	\includegraphics[width=0.19\columnwidth]{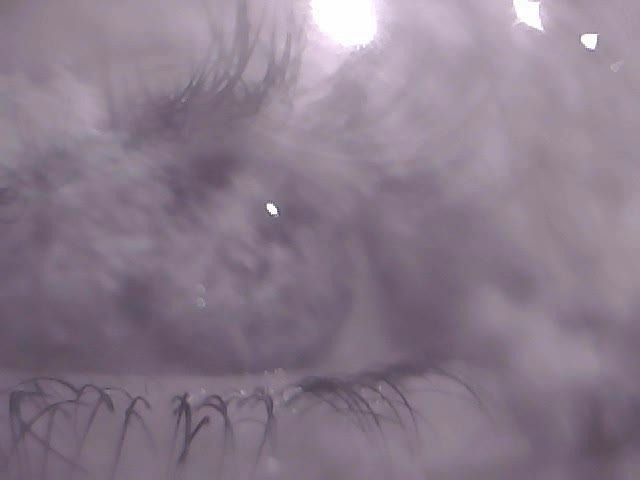}
	\includegraphics[width=0.19\columnwidth]{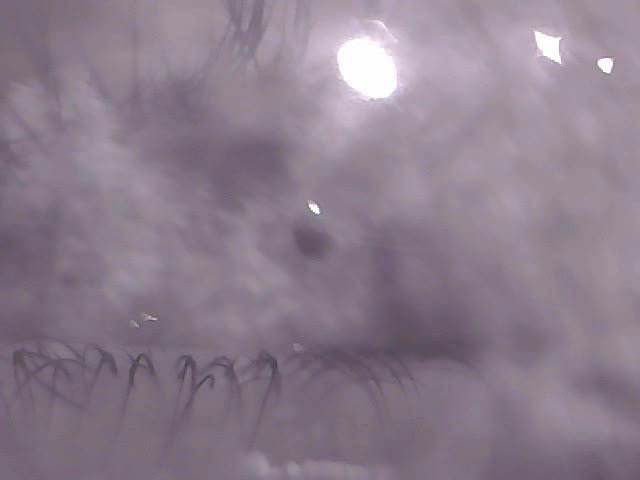}
	\includegraphics[width=0.19\columnwidth]{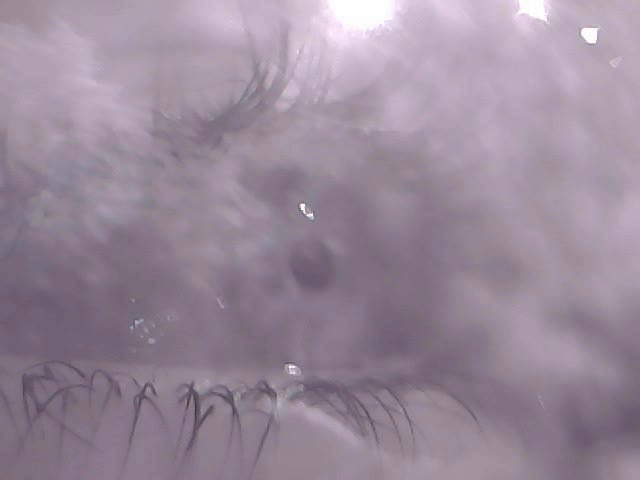}
	\includegraphics[width=0.19\columnwidth]{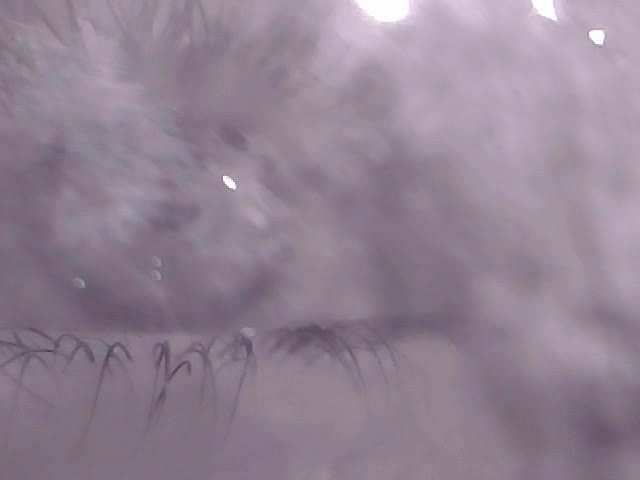}
	\includegraphics[width=0.19\columnwidth]{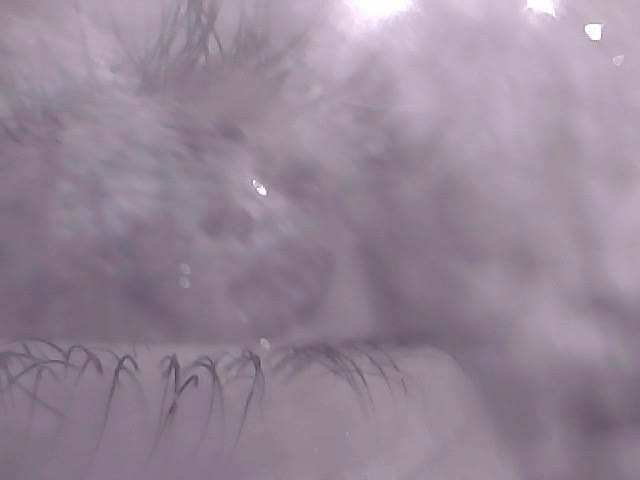}
	\caption{
		Extreme cases for pupil detection. For \ds{LPW/5/6} (top row) and
		\ds{LPW/4/1} (middle row), the eye tracker can be readjusted to improve
		detection rates. For \ds{LPW/3/16} (bottom row), readjusting the eye
		tracker is not likely to improve the conditions for pupil detection. In
		all cases, researchers should be aware that the automatic pupil
		detection is not reliable.
		\pure's confidence measure allows for users to be prompted in real time
		for adjustments and provides researchers with a quantitative metric for
		the quality of the pupil detection.
	}~\label{fig:very-hard}
\end{figure}

Moreover, there are extreme cases in which pupil detection might not be
feasible at all, such as when the bulk of the pupil is occluded due to
inadequate eye tracker placement relative to the eye.
For instance, \emph{use cases} \ds{LPW/5/6} and \ds{LPW/4/1}, for which the best
detection rates were measly $3.45\%$ (by \exc) and $14.15\%$ (by \swi),
respectively.
Sample images throughout these \emph{use cases} are shown in \figref{fig:very-hard}.
As can be seen in this figure, in the former not only the eye is
out of focus, but there are lenses obstructing mostly of the pupil, whereas in
the latter, the pupil is mostly occluded by the eyelid and eye lashes.
In such cases, \pure's confidence measure provides a quantitative measure of the
extend to which it can detect the pupil in current conditions: By observing the
ratio of confidence measures above the required threshold during a
period\footnote{The period should be significantly larger than expected blink
durations since the confidence measure is also excepted to drop during blinks;
in this section we report the ratio for the whole \emph{use case}.}.
If this ratio is too small, it can be inferred that either the pupil is not
visible or \pure{} can not cope with current conditions.
In the former case, the user can be prompted to readjust the position of the eye
tracker in real time -- this is the case for \ds{LPW/5/6}
($ratio_{th=0.66}=0.15$) and \ds{LPW/4/1} ($ratio_{th=0.66}=0.54$).
In both cases, the confidence measure ratio is useful for researchers to be
aware that the data is not reliable and requires further processing, such as
manual annotation.
An example of the cases in which adjusting the eye tracker is not likely to
improve detection rates is \emph{use case} \ds{LPW/3/16}
($ratio_{th=0.66}=0.65$), for which reflections cover most of the image as seen
in \figref{fig:very-hard}. The best detection rate for this \emph{use case} was
$31.95\%$ (by \pure).
To further support this claim, we measured the correlation between this
confidence-measure ($ratio_{th=0.66}$) and the pupil detection rate, which
resulted in a correlation coefficient of 0.88.

Envisioned future work includes investigating suitable non-edge-based second
steps, and developing a feedback system to prompt users to readjust the eye tracker
-- to be integrated into \eyerectoo\citep{santini2017eyerectoo}.
Moreover, the focus of this work is on pupil detection -- even though we
formulated the resulting signal as tracking-by-detection in \secref{sec:signal}.
Tracking methods face many challenges in eye tracking due to the fast paced
changes in illumination, frequent pupil occlusion due to blinks, and substantial
amplitude and velocity of saccadic eye movements -- as fast as
\SI{700}{\degree/\second}~\citep{baloh1975quantitative}.
Therefore, a comprehensive evaluation of tracking methods is out of the scope of
this paper and better left for future work.
Nonetheless, using temporal information (i.e., inter-frame) can greatly benefit
detection rates -- albeit care has to be taken to track the right element in the
image. In this regard, \pure's confidence metric provides a foundation that can
be used to build robust pupil trackers.


\bibliographystyle{model2-names}
\bibliography{refs}

\begin{thebibliography}{54}
\expandafter\ifx\csname natexlab\endcsname\relax\def\natexlab#1{#1}\fi
\providecommand{\url}[1]{\texttt{#1}}
\providecommand{\href}[2]{#2}
\providecommand{\path}[1]{#1}
\providecommand{\DOIprefix}{doi:}
\providecommand{\ArXivprefix}{arXiv:}
\providecommand{\URLprefix}{URL: }
\providecommand{\Pubmedprefix}{pmid:}
\providecommand{\doi}[1]{\href{http://dx.doi.org/#1}{\path{#1}}}
\providecommand{\Pubmed}[1]{\href{pmid:#1}{\path{#1}}}
\providecommand{\bibinfo}[2]{#2}
\ifx\xfnm\relax \def\xfnm[#1]{\unskip,\space#1}\fi
\bibitem[{Baloh et~al.(1975)Baloh, Sills, Kumley and
  Honrubia}]{baloh1975quantitative}
\bibinfo{author}{Baloh, R.W.}, \bibinfo{author}{Sills, A.W.},
  \bibinfo{author}{Kumley, W.E.}, \bibinfo{author}{Honrubia, V.},
  \bibinfo{year}{1975}.
\newblock \bibinfo{title}{Quantitative measurement of saccade amplitude,
  duration, and velocity}.
\newblock \bibinfo{journal}{Neurology} \bibinfo{volume}{25},
  \bibinfo{pages}{1065--1065}.
\bibitem[{Bashir and Porikli(2006)}]{bashir2006performance}
\bibinfo{author}{Bashir, F.}, \bibinfo{author}{Porikli, F.},
  \bibinfo{year}{2006}.
\newblock \bibinfo{title}{Performance evaluation of object detection and
  tracking systems}, in: \bibinfo{booktitle}{Proceedings 9th IEEE International
  Workshop on PETS}, pp. \bibinfo{pages}{7--14}.
\bibitem[{Bulling and Gellersen(2010)}]{bulling2010toward}
\bibinfo{author}{Bulling, A.}, \bibinfo{author}{Gellersen, H.},
  \bibinfo{year}{2010}.
\newblock \bibinfo{title}{Toward mobile eye-based human-computer interaction}.
\newblock \bibinfo{journal}{IEEE Pervasive Computing} \bibinfo{volume}{9},
  \bibinfo{pages}{8--12}.
\bibitem[{C.~Braunagel(2017)}]{braunagel2017ready}
\bibinfo{author}{C.~Braunagel, W.~Rosenstiel, E.K.}, \bibinfo{year}{2017}.
\newblock \bibinfo{title}{Ready for take-over? a new driver assistance system
  for an automated classification of driver take-over readiness}.
\newblock \bibinfo{journal}{IEEE Intelligent Transportation Systems Magazine} ,
  \bibinfo{pages}{in press}.
\bibitem[{Canny(1986)}]{canny1986computational}
\bibinfo{author}{Canny, J.}, \bibinfo{year}{1986}.
\newblock \bibinfo{title}{A computational approach to edge detection}.
\newblock \bibinfo{journal}{IEEE Transactions on pattern analysis and machine
  intelligence} , \bibinfo{pages}{679--698}.
\bibitem[{{\v{C}}ehovin et~al.(2016){\v{C}}ehovin, Leonardis and
  Kristan}]{vcehovin2016visual}
\bibinfo{author}{{\v{C}}ehovin, L.}, \bibinfo{author}{Leonardis, A.},
  \bibinfo{author}{Kristan, M.}, \bibinfo{year}{2016}.
\newblock \bibinfo{title}{Visual object tracking performance measures
  revisited}.
\newblock \bibinfo{journal}{IEEE Transactions on Image Processing}
  \bibinfo{volume}{25}, \bibinfo{pages}{1261--1274}.
\bibitem[{Chetlur et~al.(2014)Chetlur, Woolley, Vandermersch, Cohen, Tran,
  Catanzaro and Shelhamer}]{chetlur2014cudnn}
\bibinfo{author}{Chetlur, S.}, \bibinfo{author}{Woolley, C.},
  \bibinfo{author}{Vandermersch, P.}, \bibinfo{author}{Cohen, J.},
  \bibinfo{author}{Tran, J.}, \bibinfo{author}{Catanzaro, B.},
  \bibinfo{author}{Shelhamer, E.}, \bibinfo{year}{2014}.
\newblock \bibinfo{title}{cudnn: Efficient primitives for deep learning}.
\newblock \bibinfo{journal}{arXiv preprint arXiv:1410.0759} .
\bibitem[{Chu et~al.(2010)Chu, Wood and Collins}]{chu2010effect}
\bibinfo{author}{Chu, B.S.}, \bibinfo{author}{Wood, J.M.},
  \bibinfo{author}{Collins, M.J.}, \bibinfo{year}{2010}.
\newblock \bibinfo{title}{The effect of presbyopic vision corrections on
  nighttime driving performance}.
\newblock \bibinfo{journal}{Investigative ophthalmology \& visual science}
  \bibinfo{volume}{51}, \bibinfo{pages}{4861--4866}.
\bibitem[{Efland et~al.(2016)Efland, Parikh, Sanghavi and
  Farooqui}]{efland2016high}
\bibinfo{author}{Efland, G.}, \bibinfo{author}{Parikh, S.},
  \bibinfo{author}{Sanghavi, H.}, \bibinfo{author}{Farooqui, A.},
  \bibinfo{year}{2016}.
\newblock \bibinfo{title}{High performance dsp for vision, imaging and neural
  networks}.
\newblock \bibinfo{journal}{IEEE Hot Chips} \bibinfo{volume}{28}.
\bibitem[{{Ergoneers}(2017)}]{ergoneers2017}
\bibinfo{author}{{Ergoneers}}, \bibinfo{year}{2017}.
\newblock \bibinfo{title}{{Dikablis Glasses Professional}}.
\newblock \bibinfo{howpublished}{Accessed in 2017-07-26}.
\newblock \URLprefix \url{http://www.ergoneers.com/en/hardware/eye-tracking/}.
\bibitem[{Fitzgibbon and Fisher(1995)}]{fitzgibbon1995buyer}
\bibinfo{author}{Fitzgibbon, A.W.}, \bibinfo{author}{Fisher, R.B.},
  \bibinfo{year}{1995}.
\newblock \bibinfo{title}{A buyer's guide to conic fitting}, in:
  \bibinfo{booktitle}{Proceedings of the 6th British Conference on Machine
  Vision (Vol. 2)}, \bibinfo{publisher}{BMVA Press}, \bibinfo{address}{Surrey,
  UK, UK}. pp. \bibinfo{pages}{513--522}.
\newblock \URLprefix \url{http://dl.acm.org/citation.cfm?id=243124.243148}.
\bibitem[{Foulsham et~al.(2011)Foulsham, Walker and
  Kingstone}]{foulsham2011and}
\bibinfo{author}{Foulsham, T.}, \bibinfo{author}{Walker, E.},
  \bibinfo{author}{Kingstone, A.}, \bibinfo{year}{2011}.
\newblock \bibinfo{title}{The where, what and when of gaze allocation in the
  lab and the natural environment}.
\newblock \bibinfo{journal}{Vision research} \bibinfo{volume}{51},
  \bibinfo{pages}{1920--1931}.
\bibitem[{Frigge et~al.(1989)Frigge, Hoaglin and Iglewicz}]{frigge1989some}
\bibinfo{author}{Frigge, M.}, \bibinfo{author}{Hoaglin, D.C.},
  \bibinfo{author}{Iglewicz, B.}, \bibinfo{year}{1989}.
\newblock \bibinfo{title}{Some implementations of the boxplot}.
\newblock \bibinfo{journal}{The American Statistician} \bibinfo{volume}{43},
  \bibinfo{pages}{50--54}.
\bibitem[{Fuhl et~al.(2016a)Fuhl, Geisler, Santini, Rosenstiel and
  Kasneci}]{fuhl2016evaluation}
\bibinfo{author}{Fuhl, W.}, \bibinfo{author}{Geisler, D.},
  \bibinfo{author}{Santini, T.}, \bibinfo{author}{Rosenstiel, W.},
  \bibinfo{author}{Kasneci, E.}, \bibinfo{year}{2016}a.
\newblock \bibinfo{title}{Evaluation of state-of-the-art pupil detection
  algorithms on remote eye images}, in: \bibinfo{booktitle}{Proceedings of the
  2016 ACM International Joint Conference on Pervasive and Ubiquitous
  Computing: Adjunct}, \bibinfo{organization}{ACM}. pp.
  \bibinfo{pages}{1716--1725}.
\bibitem[{Fuhl et~al.(2015)Fuhl, K{\"u}bler, Sippel, Rosenstiel and
  Kasneci}]{fuhl2015excuse}
\bibinfo{author}{Fuhl, W.}, \bibinfo{author}{K{\"u}bler, T.},
  \bibinfo{author}{Sippel, K.}, \bibinfo{author}{Rosenstiel, W.},
  \bibinfo{author}{Kasneci, E.}, \bibinfo{year}{2015}.
\newblock \bibinfo{title}{Excuse: Robust pupil detection in real-world
  scenarios}, in: \bibinfo{booktitle}{International Conference on Computer
  Analysis of Images and Patterns}, \bibinfo{organization}{Springer}. pp.
  \bibinfo{pages}{39--51}.
\bibitem[{Fuhl et~al.(2016b)Fuhl, Santini, Kasneci and
  Kasneci}]{fuhl2016pupilnet}
\bibinfo{author}{Fuhl, W.}, \bibinfo{author}{Santini, T.},
  \bibinfo{author}{Kasneci, G.}, \bibinfo{author}{Kasneci, E.},
  \bibinfo{year}{2016}b.
\newblock \bibinfo{title}{Pupilnet: convolutional neural networks for robust
  pupil detection}.
\newblock \bibinfo{journal}{arXiv preprint arXiv:1601.04902} .
\bibitem[{Fuhl et~al.(2016c)Fuhl, Santini, K{\"u}bler and
  Kasneci}]{fuhl2016else}
\bibinfo{author}{Fuhl, W.}, \bibinfo{author}{Santini, T.C.},
  \bibinfo{author}{K{\"u}bler, T.}, \bibinfo{author}{Kasneci, E.},
  \bibinfo{year}{2016}c.
\newblock \bibinfo{title}{Else: Ellipse selection for robust pupil detection in
  real-world environments}, in: \bibinfo{booktitle}{Proceedings of the Ninth
  Biennial ACM Symposium on Eye Tracking Research \& Applications},
  \bibinfo{organization}{ACM}. pp. \bibinfo{pages}{123--130}.
\bibitem[{Fuhl et~al.(2016d)Fuhl, Tonsen, Bulling and Kasneci}]{fuhl2016pupil}
\bibinfo{author}{Fuhl, W.}, \bibinfo{author}{Tonsen, M.},
  \bibinfo{author}{Bulling, A.}, \bibinfo{author}{Kasneci, E.},
  \bibinfo{year}{2016}d.
\newblock \bibinfo{title}{Pupil detection for head-mounted eye tracking in the
  wild: an evaluation of the state of the art}.
\newblock \bibinfo{journal}{Machine Vision and Applications}
  \bibinfo{volume}{27}, \bibinfo{pages}{1275--1288}.
\bibitem[{Guenter et~al.(2012)Guenter, Finch, Drucker, Tan and
  Snyder}]{guenter2012foveated}
\bibinfo{author}{Guenter, B.}, \bibinfo{author}{Finch, M.},
  \bibinfo{author}{Drucker, S.}, \bibinfo{author}{Tan, D.},
  \bibinfo{author}{Snyder, J.}, \bibinfo{year}{2012}.
\newblock \bibinfo{title}{Foveated 3d graphics}.
\newblock \bibinfo{journal}{ACM Transactions on Graphics (TOG)}
  \bibinfo{volume}{31}, \bibinfo{pages}{164}.
\bibitem[{Hansen and Hammoud(2007)}]{hansen2007improved}
\bibinfo{author}{Hansen, D.W.}, \bibinfo{author}{Hammoud, R.I.},
  \bibinfo{year}{2007}.
\newblock \bibinfo{title}{An improved likelihood model for eye tracking}.
\newblock \bibinfo{journal}{Computer Vision and Image Understanding}
  \bibinfo{volume}{106}, \bibinfo{pages}{220--230}.
\bibitem[{Hansen and Pece(2005)}]{hansen2005eye}
\bibinfo{author}{Hansen, D.W.}, \bibinfo{author}{Pece, A.E.},
  \bibinfo{year}{2005}.
\newblock \bibinfo{title}{Eye tracking in the wild}.
\newblock \bibinfo{journal}{Computer Vision and Image Understanding}
  \bibinfo{volume}{98}, \bibinfo{pages}{155--181}.
\bibitem[{Jansen et~al.(2009)Jansen, Kingma and Peeters}]{jansen2009confidence}
\bibinfo{author}{Jansen, S.}, \bibinfo{author}{Kingma, H.},
  \bibinfo{author}{Peeters, R.}, \bibinfo{year}{2009}.
\newblock \bibinfo{title}{A confidence measure for real-time eye movement
  detection in video-oculography}, in: \bibinfo{booktitle}{13th International
  Conference on Biomedical Engineering}, \bibinfo{organization}{Springer}. pp.
  \bibinfo{pages}{335--339}.
\bibitem[{Kasneci(2013)}]{kasneci2013towards}
\bibinfo{author}{Kasneci, E.}, \bibinfo{year}{2013}.
\newblock \bibinfo{title}{Towards the automated recognition of assistance need
  for drivers with impaired visual field}.
\newblock Ph.D. thesis. Universit{\"a}t T{\"u}bingen, Germany.
\bibitem[{Kasneci et~al.(2014)Kasneci, Sippel, Heister, Aehling, Rosenstiel,
  Schiefer and Papageorgiou}]{kasneci2014homonymous}
\bibinfo{author}{Kasneci, E.}, \bibinfo{author}{Sippel, K.},
  \bibinfo{author}{Heister, M.}, \bibinfo{author}{Aehling, K.},
  \bibinfo{author}{Rosenstiel, W.}, \bibinfo{author}{Schiefer, U.},
  \bibinfo{author}{Papageorgiou, E.}, \bibinfo{year}{2014}.
\newblock \bibinfo{title}{Homonymous visual field loss and its impact on visual
  exploration: A supermarket study}.
\newblock \bibinfo{journal}{Translational vision science \& technology}
  \bibinfo{volume}{3}, \bibinfo{pages}{2--2}.
\bibitem[{Kristan et~al.(2016)Kristan, Matas, Leonardis, Voj{\'\i}{\v{r}},
  Pflugfelder, Fernandez, Nebehay, Porikli and
  {\v{C}}ehovin}]{kristan2016novel}
\bibinfo{author}{Kristan, M.}, \bibinfo{author}{Matas, J.},
  \bibinfo{author}{Leonardis, A.}, \bibinfo{author}{Voj{\'\i}{\v{r}}, T.},
  \bibinfo{author}{Pflugfelder, R.}, \bibinfo{author}{Fernandez, G.},
  \bibinfo{author}{Nebehay, G.}, \bibinfo{author}{Porikli, F.},
  \bibinfo{author}{{\v{C}}ehovin, L.}, \bibinfo{year}{2016}.
\newblock \bibinfo{title}{A novel performance evaluation methodology for
  single-target trackers}.
\newblock \bibinfo{journal}{IEEE transactions on pattern analysis and machine
  intelligence} \bibinfo{volume}{38}, \bibinfo{pages}{2137--2155}.
\bibitem[{K{\"u}bler et~al.(2015)K{\"u}bler, Kasneci, Rosenstiel, Heister,
  Aehling, Nagel, Schiefer and Papageorgiou}]{kubler2015driving}
\bibinfo{author}{K{\"u}bler, T.C.}, \bibinfo{author}{Kasneci, E.},
  \bibinfo{author}{Rosenstiel, W.}, \bibinfo{author}{Heister, M.},
  \bibinfo{author}{Aehling, K.}, \bibinfo{author}{Nagel, K.},
  \bibinfo{author}{Schiefer, U.}, \bibinfo{author}{Papageorgiou, E.},
  \bibinfo{year}{2015}.
\newblock \bibinfo{title}{Driving with glaucoma: task performance and gaze
  movements}.
\newblock \bibinfo{journal}{Optometry \& Vision Science} \bibinfo{volume}{92},
  \bibinfo{pages}{1037--1046}.
\bibitem[{Kunjur et~al.(2006)Kunjur, Sabesan and
  Ilankovan}]{kunjur2006anthropometric}
\bibinfo{author}{Kunjur, J.}, \bibinfo{author}{Sabesan, T.},
  \bibinfo{author}{Ilankovan, V.}, \bibinfo{year}{2006}.
\newblock \bibinfo{title}{Anthropometric analysis of eyebrows and eyelids: an
  inter-racial study}.
\newblock \bibinfo{journal}{British Journal of Oral and Maxillofacial Surgery}
  \bibinfo{volume}{44}, \bibinfo{pages}{89--93}.
\bibitem[{Liu et~al.(2002)Liu, Xu and Fujimura}]{liu2002real}
\bibinfo{author}{Liu, X.}, \bibinfo{author}{Xu, F.}, \bibinfo{author}{Fujimura,
  K.}, \bibinfo{year}{2002}.
\newblock \bibinfo{title}{Real-time eye detection and tracking for driver
  observation under various light conditions}, in:
  \bibinfo{booktitle}{Intelligent Vehicle Symposium, 2002. IEEE}.
\bibitem[{Microsoft(2017)}]{microsoft2017}
\bibinfo{author}{Microsoft}, \bibinfo{year}{2017}.
\newblock \bibinfo{howpublished}{Accessed in 2017-07-26}.
\newblock \URLprefix \url{https://www.microsoft.com/en-us/hololens}.
\bibitem[{Mohammed et~al.(2012)Mohammed, Hong and
  Jarjes}]{mohammed2012accurate}
\bibinfo{author}{Mohammed, G.J.}, \bibinfo{author}{Hong, B.R.},
  \bibinfo{author}{Jarjes, A.A.}, \bibinfo{year}{2012}.
\newblock \bibinfo{title}{Accurate pupil features extraction based on new
  projection function}.
\newblock \bibinfo{journal}{Computing and Informatics} \bibinfo{volume}{29},
  \bibinfo{pages}{663--680}.
\bibitem[{Morimoto and Mimica(2005)}]{morimoto2005eye}
\bibinfo{author}{Morimoto, C.H.}, \bibinfo{author}{Mimica, M.R.},
  \bibinfo{year}{2005}.
\newblock \bibinfo{title}{Eye gaze tracking techniques for interactive
  applications}.
\newblock \bibinfo{journal}{Computer vision and image understanding}
  \bibinfo{volume}{98}, \bibinfo{pages}{4--24}.
\bibitem[{Oculus(2017)}]{oculus2017}
\bibinfo{author}{Oculus}, \bibinfo{year}{2017}.
\newblock \bibinfo{howpublished}{Accessed in 2017-07-26}.
\newblock \URLprefix \url{https://www.oculus.com/rift/}.
\bibitem[{Pedrotti et~al.(2011)Pedrotti, Lei, Dzaack and
  R{\"o}tting}]{pedrotti2011data}
\bibinfo{author}{Pedrotti, M.}, \bibinfo{author}{Lei, S.},
  \bibinfo{author}{Dzaack, J.}, \bibinfo{author}{R{\"o}tting, M.},
  \bibinfo{year}{2011}.
\newblock \bibinfo{title}{A data-driven algorithm for offline pupil signal
  preprocessing and eyeblink detection in low-speed eye-tracking protocols}.
\newblock \bibinfo{journal}{Behavior Research Methods} \bibinfo{volume}{43},
  \bibinfo{pages}{372--383}.
\bibitem[{Pheatt(2008)}]{pheatt2008intel}
\bibinfo{author}{Pheatt, C.}, \bibinfo{year}{2008}.
\newblock \bibinfo{title}{Intel{\textregistered} threading building blocks}.
\newblock \bibinfo{journal}{Journal of Computing Sciences in Colleges}
  \bibinfo{volume}{23}, \bibinfo{pages}{298--298}.
\bibitem[{{Pupil Labs}(2017)}]{pupillabs2017}
\bibinfo{author}{{Pupil Labs}}, \bibinfo{year}{2017}.
\newblock \bibinfo{howpublished}{Accessed in 2017-07-26}.
\newblock \URLprefix \url{https://pupil-labs.com/}.
\bibitem[{Raffle and Wang(2015)}]{raffle2015heads}
\bibinfo{author}{Raffle, H.S.}, \bibinfo{author}{Wang, C.J.},
  \bibinfo{year}{2015}.
\newblock \bibinfo{title}{Heads up display}.
\newblock \bibinfo{note}{US Patent 9,001,030}.
\bibitem[{Santini et~al.(2017a)Santini, Fuhl, Geisler and
  Kasneci}]{santini2017eyerectoo}
\bibinfo{author}{Santini, T.}, \bibinfo{author}{Fuhl, W.},
  \bibinfo{author}{Geisler, D.}, \bibinfo{author}{Kasneci, E.},
  \bibinfo{year}{2017}a.
\newblock \bibinfo{title}{Eyerectoo: Open-source software for real-time
  pervasive head-mounted eye-tracking}, in: \bibinfo{booktitle}{Proceedings of
  the 12th Joint Conference on Computer Vision, Imaging and Computer Graphics
  Theory and Applications}.
\bibitem[{Santini et~al.(2017b)Santini, Fuhl and Kasneci}]{santini2017calibme}
\bibinfo{author}{Santini, T.}, \bibinfo{author}{Fuhl, W.},
  \bibinfo{author}{Kasneci, E.}, \bibinfo{year}{2017}b.
\newblock \bibinfo{title}{Calibme: Fast and unsupervised eye tracker
  calibration for gaze-based pervasive human-computer interaction}, in:
  \bibinfo{booktitle}{Proceedings of the 2017 CHI Conference on Human Factors
  in Computing Systems}, \bibinfo{organization}{ACM}. pp.
  \bibinfo{pages}{2594--2605}.
\bibitem[{Santini et~al.(2016)Santini, Fuhl, K{\"u}bler and
  Kasneci}]{santini2016bayesian}
\bibinfo{author}{Santini, T.}, \bibinfo{author}{Fuhl, W.},
  \bibinfo{author}{K{\"u}bler, T.}, \bibinfo{author}{Kasneci, E.},
  \bibinfo{year}{2016}.
\newblock \bibinfo{title}{Bayesian identification of fixations, saccades, and
  smooth pursuits}, in: \bibinfo{booktitle}{Proceedings of the Ninth Biennial
  ACM Symposium on Eye Tracking Research \& Applications},
  \bibinfo{organization}{ACM}. pp. \bibinfo{pages}{163--170}.
\bibitem[{Schmidt et~al.(2017)Schmidt, Laarousi, Stolzmann and
  Karrer-Gau{\ss}}]{schmidt2017eye}
\bibinfo{author}{Schmidt, J.}, \bibinfo{author}{Laarousi, R.},
  \bibinfo{author}{Stolzmann, W.}, \bibinfo{author}{Karrer-Gau{\ss}, K.},
  \bibinfo{year}{2017}.
\newblock \bibinfo{title}{Eye blink detection for different driver states in
  conditionally automated driving and manual driving using eog and a driver
  camera}.
\newblock \bibinfo{journal}{Behavior Research Methods} ,
  \bibinfo{pages}{1--14}.
\bibitem[{Spector(1990)}]{spector1990clinical}
\bibinfo{author}{Spector, R.}, \bibinfo{year}{1990}.
\newblock \bibinfo{title}{The pupils}, in: \bibinfo{editor}{Walker~HK, Hall~WD,
  H.J.} (Ed.), \bibinfo{booktitle}{Clinical Methods: The HIstory, Physical, and
  Laboratory Examinations}. \bibinfo{publisher}{Butterworths}.
  chapter~\bibinfo{chapter}{8}.
\bibitem[{Sugano and Bulling(2015)}]{sugano2015self}
\bibinfo{author}{Sugano, Y.}, \bibinfo{author}{Bulling, A.},
  \bibinfo{year}{2015}.
\newblock \bibinfo{title}{Self-calibrating head-mounted eye trackers using
  egocentric visual saliency}, in: \bibinfo{booktitle}{Proceedings of the 28th
  Annual ACM Symposium on User Interface Software \& Technology},
  \bibinfo{organization}{ACM}. pp. \bibinfo{pages}{363--372}.
\bibitem[{{\'S}wirski et~al.(2012){\'S}wirski, Bulling and
  Dodgson}]{swirski2012robust}
\bibinfo{author}{{\'S}wirski, L.}, \bibinfo{author}{Bulling, A.},
  \bibinfo{author}{Dodgson, N.}, \bibinfo{year}{2012}.
\newblock \bibinfo{title}{Robust real-time pupil tracking in highly off-axis
  images}, in: \bibinfo{booktitle}{Proceedings of the Symposium on Eye Tracking
  Research and Applications}, \bibinfo{organization}{ACM}. pp.
  \bibinfo{pages}{173--176}.
\bibitem[{\'Swirski and Dodgson(2013)}]{swirski2013fully}
\bibinfo{author}{\'Swirski, L.}, \bibinfo{author}{Dodgson, N.A.},
  \bibinfo{year}{2013}.
\newblock \bibinfo{title}{A fully-automatic, temporal approach to single
  camera, glint-free 3d eye model fitting [abstract]}, in:
  \bibinfo{booktitle}{Proceedings of ECEM 2013}.
\bibitem[{Teh and Chin(1989)}]{teh1989detection}
\bibinfo{author}{Teh, C.H.}, \bibinfo{author}{Chin, R.T.},
  \bibinfo{year}{1989}.
\newblock \bibinfo{title}{On the detection of dominant points on digital
  curves}.
\newblock \bibinfo{journal}{IEEE Transactions on pattern analysis and machine
  intelligence} \bibinfo{volume}{11}, \bibinfo{pages}{859--872}.
\bibitem[{Tien et~al.(2015)Tien, Pucher, Sodergren, Sriskandarajah, Yang and
  Darzi}]{tien2015differences}
\bibinfo{author}{Tien, T.}, \bibinfo{author}{Pucher, P.H.},
  \bibinfo{author}{Sodergren, M.H.}, \bibinfo{author}{Sriskandarajah, K.},
  \bibinfo{author}{Yang, G.Z.}, \bibinfo{author}{Darzi, A.},
  \bibinfo{year}{2015}.
\newblock \bibinfo{title}{Differences in gaze behaviour of expert and junior
  surgeons performing open inguinal hernia repair}.
\newblock \bibinfo{journal}{Surgical endoscopy} \bibinfo{volume}{29},
  \bibinfo{pages}{405--413}.
\bibitem[{Tonsen et~al.(2016)Tonsen, Zhang, Sugano and
  Bulling}]{tonsen2016labelled}
\bibinfo{author}{Tonsen, M.}, \bibinfo{author}{Zhang, X.},
  \bibinfo{author}{Sugano, Y.}, \bibinfo{author}{Bulling, A.},
  \bibinfo{year}{2016}.
\newblock \bibinfo{title}{Labelled pupils in the wild: a dataset for studying
  pupil detection in unconstrained environments}, in:
  \bibinfo{booktitle}{Proceedings of the Ninth Biennial ACM Symposium on Eye
  Tracking Research \& Applications}, \bibinfo{organization}{ACM}. pp.
  \bibinfo{pages}{139--142}.
\bibitem[{Toussaint(1983)}]{toussaint1983solving}
\bibinfo{author}{Toussaint, G.T.}, \bibinfo{year}{1983}.
\newblock \bibinfo{title}{Solving geometric problems with the rotating
  calipers}, in: \bibinfo{booktitle}{Proc. IEEE Melecon}, p.
  \bibinfo{pages}{A10}.
\bibitem[{Tr{\"o}sterer et~al.(2014)Tr{\"o}sterer, Meschtscherjakov, Wilfinger
  and Tscheligi}]{trosterer2014eye}
\bibinfo{author}{Tr{\"o}sterer, S.}, \bibinfo{author}{Meschtscherjakov, A.},
  \bibinfo{author}{Wilfinger, D.}, \bibinfo{author}{Tscheligi, M.},
  \bibinfo{year}{2014}.
\newblock \bibinfo{title}{Eye tracking in the car: Challenges in a dual-task
  scenario on a test track}, in: \bibinfo{booktitle}{Proceedings of the 6th
  AutomotiveUI}, \bibinfo{organization}{ACM}.
\bibitem[{Vera-Olmos and Malpica(2017)}]{vera2017deconvolutional}
\bibinfo{author}{Vera-Olmos, F.}, \bibinfo{author}{Malpica, N.},
  \bibinfo{year}{2017}.
\newblock \bibinfo{title}{Deconvolutional neural network for pupil detection in
  real-world environments}, in: \bibinfo{booktitle}{International
  Work-Conference on the Interplay Between Natural and Artificial Computation},
  \bibinfo{organization}{Springer}. pp. \bibinfo{pages}{223--231}.
\bibitem[{Vidal et~al.(2012)Vidal, Turner, Bulling and
  Gellersen}]{vidal2012wearable}
\bibinfo{author}{Vidal, M.}, \bibinfo{author}{Turner, J.},
  \bibinfo{author}{Bulling, A.}, \bibinfo{author}{Gellersen, H.},
  \bibinfo{year}{2012}.
\newblock \bibinfo{title}{Wearable eye tracking for mental health monitoring}.
\newblock \bibinfo{journal}{Computer Communications} \bibinfo{volume}{35},
  \bibinfo{pages}{1306--1311}.
\bibitem[{Vrzakova and Bednarik(2012)}]{vrzakova2012hard}
\bibinfo{author}{Vrzakova, H.}, \bibinfo{author}{Bednarik, R.},
  \bibinfo{year}{2012}.
\newblock \bibinfo{title}{Hard lessons learned: mobile eye-tracking in
  cockpits}, in: \bibinfo{booktitle}{Proceedings of the 4th Workshop on Eye
  Gaze in Intelligent Human Machine Interaction}, \bibinfo{organization}{ACM}.
  p.~\bibinfo{pages}{7}.
\bibitem[{Wood et~al.(2017)Wood, Tyrrell, Lacherez and Black}]{wood2017night}
\bibinfo{author}{Wood, J.M.}, \bibinfo{author}{Tyrrell, R.A.},
  \bibinfo{author}{Lacherez, P.}, \bibinfo{author}{Black, A.A.},
  \bibinfo{year}{2017}.
\newblock \bibinfo{title}{Night-time pedestrian conspicuity: effects of
  clothing on drivers’ eye movements}.
\newblock \bibinfo{journal}{Ophthalmic and physiological optics}
  \bibinfo{volume}{37}, \bibinfo{pages}{184--190}.
\bibitem[{Zhu and Ji(2005)}]{zhu2005robust}
\bibinfo{author}{Zhu, Z.}, \bibinfo{author}{Ji, Q.}, \bibinfo{year}{2005}.
\newblock \bibinfo{title}{Robust real-time eye detection and tracking under
  variable lighting conditions and various face orientations}.
\newblock \bibinfo{journal}{Computer Vision and Image Understanding}
  \bibinfo{volume}{98}, \bibinfo{pages}{124--154}.

\end{thebibliography}

\end{document}